\documentclass[11pt,a4paper]{article}
\pdfoutput=0

\usepackage[margin=.9in]{geometry}
\usepackage{authblk}
\usepackage{verbatim}
\usepackage{amsmath}
\usepackage{amsthm}
\usepackage{amsfonts}
\usepackage{amssymb}
\usepackage{algorithm2e}
\usepackage{graphicx}
\usepackage{subcaption}
\usepackage{bbm}
\usepackage{multicol,color}
\usepackage{enumitem}
\usepackage[round]{natbib}
\usepackage{hyperref}

\def\equationautorefname~#1\null{Equation~(#1)\null}
\def\algorithmautorefname~#1\null{Algorithm~#1\null}

\newcommand*{\sref}[1]{\hyperref[{#1}]{Section~\ref*{#1}}}
\newcommand*{\pref}[1]{\hyperref[{#1}]{Problem~(\ref*{#1})}}
\newcommand*{\fref}[2]{\hyperref[{#1}]{\autoref*{#1}~(#2)}}
\newcommand*{\frefs}[3]{\hyperref[{#1}]{\autoref*{#1}~(#2-#3)}}
\newcommand*{\app}[1]{\hyperref[{#1}]{Appendix}}

\renewcommand{\H}{\mathbf{H}}

\newcommand{\R}{\mathbb{R}}
\newcommand{\V}{\mathbf{V}}
\newcommand{\W}{\mathbf{W}}
\newcommand{\X}{\mathbf{X}}
\newcommand{\Y}{\mathbf{Y}}
\newcommand{\Z}{\mathbf{Z}}
\newcommand{\prox}{\mathbf{prox}}
\newcommand{\1}{\mathbf{1}}
\newcommand{\T}{\top}

\begin{document}

\title{Primal-Dual Algorithms for Non-negative Matrix Factorization with the Kullback-Leibler Divergence}

\author{Felipe Yanez}
\author{Francis Bach}

\affil{INRIA -- SIERRA Project-Team\\ D\'epartement d'Informatique de l'\'Ecole Normale Sup\'erieure\\ Paris, France}

\renewcommand\Affilfont{\itshape\small}
\renewcommand\Authand{ and }

\maketitle


\begin{abstract}
Non-negative matrix factorization (NMF) approximates a given matrix as a product of two non-negative matrices. Multiplicative algorithms deliver reliable results, but they show slow convergence for high-dimensional data and may be stuck away from local minima.
    	Gradient descent methods have better behavior, but only apply to smooth losses such as the least-squares loss. 
    	In this article, we propose a first-order primal-dual algorithm for non-negative decomposition problems (where one factor is fixed) with the KL divergence, based on the Chambolle-Pock algorithm.  All required computations may be obtained in closed form and we provide an efficient heuristic way to select step-sizes. By using alternating optimization, our algorithm readily extends to NMF and, on synthetic examples, face recognition or music source separation datasets, it is either faster than existing algorithms, or leads to improved local optima, or both.
\end{abstract}


\section{Introduction}

The current development of techniques for big data applications has been extremely useful in many fields including data analysis, bioinformatics and scientific computing. These techniques need to handle large amounts of data and often rely on dimensionality reduction; this is often cast as approximating a matrix with a low-rank element.\\

Non-negative matrix factorization (NMF) is a method that aims at finding part-based, linear representations of non-negative data by factorizing it as the product of two low-rank non-negative matrices~\citep{paatero1994,lee1999}. In \citeyear{lee2000}, two multiplicative algorithms for NMF were introduced by \citeauthor{lee2000}, one that minimizes the conventional least-squares error, and other one that minimizes the generalized Kullback-Leibler (KL) divergence~\citep{lee2000}.\\
 
These algorithms extend to other losses and have been reported in different applications, e.g., face recognition \citep{wang2005}, music analysis~\citep{fevotte2009a}, and text mining \citep{guduru2006}. An important weakness of multiplicative algorithms is their slow convergence rate in high-dimensional data and their susceptibility to become trapped in poor local optima~\citep{lin2007}. Gradient descent methods for NMF provide additional flexibility and fast convergence~\citep{lin2007,kim2008,gillis2011}. These methods have been extensively studied for the minimization of the least-squares error~\citep{lin2007,kim2008}.\\

The goal of this paper is to provide similar first-order methods for the KL divergence, with  updates as cheap as multiplicative updates. Our method builds on the recent work of \citet{sun2014} which consider the alternating direction method of multipliers (ADMM) adapted to this problem. We instead rely on the Chambolle-Pock algorithm~\citep{chambolle2011}, which may be seen as a linearized version of ADMM, and thus we may reuse some of the tools developed by~\citet{sun2014} while having an empirically faster algorithm.


\subsection{Contributions}

The main contributions of this article are as follows: 

\begin{itemize}
\item[--] We propose a new primal-dual formulation for the convex KL decomposition problem in \sref{sec:pd}, and an extension to the non-convex problem of NMF by alternating minimization in \sref{sec:implementation}.

\item[--] We provide a purely data-driven way to select all step-sizes of our algorithm in \sref{sec:stepsizes}.

\item[--] In our simulations in \sref{sec:results} on synthetic examples, face recognition or music source separation datasets, our algorithm is either faster than existing algorithms, or leads to improved local optima, or both.

\item[--] We derive a cheap and efficient implementation (\autoref{algo:2}).~Matlab code is available online at: \texttt{anonymized website} 
\end{itemize}


\section{Problem Formulation}

Let $\V\in\R^{n\times m}_+$ denote the $n\times m$ given matrix formed by $m$ non-negative column vectors of dimensionality~$n$. Considering $r\leq\min(n,m)$,  let $\W\in\R^{m\times r}_+$ and $\H\in\R^{r\times n}_+$ be the matrix factors such that

\begin{equation*}
\V\approx\W\H.
\end{equation*}

Two widely used cost functions for NMF are the conventional least-squares error (not detailed herein), and the generalized KL divergence

\begin{eqnarray}
D(\V\|\W\H) &=& - \sum_{i=1}^m\sum_{j=1}^n\V_{ij}\left\{\log\left(\frac{(\W\H)_{ij}}{\V_{ij}}\right) + 1\right\} \nonumber \\
&& \qquad \ \ \ \ \ + \sum_{i=1}^m\sum_{j=1}^n(\W\H)_{ij}.
\label{eq:KLdivergence}
\end{eqnarray}

In this work, only the KL divergence is considered. Therefore, the optimization problem is as follows:

\begin{equation}
\begin{aligned}
& \underset{\W,\H\;\geq\;0}{\operatorname{minimize}}
& & D(\V\|\W\H).
\end{aligned}
\label{prob:nmf}
\end{equation}

We recall that the previous problem is non-convex in both factors simultaneously, whereas convex in each factor separately, i.e., the non-negative  decomposition (ND) problems,

\begin{eqnarray}
& &\underset{\W\;\geq\;0}{\operatorname{minimize}}\;D(\V\|\W\H) \label{prob:nd1}\\
&\mathrm{and} &\underset{\H\;\geq\;0}{\operatorname{minimize}}\;D(\V\|\W\H),\label{prob:nd2}
\end{eqnarray}

are convex.\\


We now present two algorithms for NMF, multiplicative updates~\citep{lee2000}, and the ADMM-based approach \citep{sun2014}. 


\subsection{Multiplicative updates}

\citet{lee2000} introduced two multiplicative updates algorithms for NMF. One minimizes the conventional least-squares error, and the other one minimizes the KL divergence.\\

The NMF problem, for both losses, is a non-convex problem in $\W$ and $\H$ simultaneously, but convex with respect to each variable taken separately; this make alternating optimization techniques, i.e., solving at each iteration two separate convex problems, very adapted: first fixing $\H$ to estimate $\W$, and then fixing $\W$ to estimate $\H$ \citep{lee2000,fevotte2009a}. The multiplicative updates algorithms (like ours) follow this approach.\\

For the KL divergence loss, the multiplicative update rule~\citep{lee2000} for $\W$ and $\H$ is as follows and may be derived from expectation-maximization (EM) for a certain probabilistic model~\citep{lee2000,fevotte2009b}:

\begin{eqnarray*}
\W_{ia} &\gets& \W_{ia}\frac{\sum_{\mu=1}^n\H_{a\mu}\V_{i\mu}/(\W\H)_{i\mu}}{\sum_{\nu=1}^n\H_{a\nu}},\;\mathrm{and} \\
\H_{a\mu} &\gets& \H_{a\mu}\frac{\sum_{i=1}^m\W_{ia}\V_{i\mu}/(\W\H)_{i\mu}}{\sum_{k=1}^m\W_{ka}}.
\label{eq:MU}
\end{eqnarray*}
 
 The complexity per iteration is $O(rmn)$.
 

\subsection{Alternating direction method of multipliers (ADMM)}

\citet{sun2014} propose an ADMM technique to solve \pref{prob:nmf}  by  reformulating it as

\begin{equation*}
\begin{aligned}
& \text{minimize}
& & D(\V\|\X)\\
& \text{subject to}
& & \X = \Y\Z\\
&
&& \Y = \W,\; \Z = \H\\
&
&& \W \geq 0,\; \H \geq 0.\\
\end{aligned}
\end{equation*}

The  updates for the primal variables $\W$, $\H$, $\X$, $\Y$ and $\Z$ are as follows and involve certain proximal operators for the KL loss which are the same as ours in \sref{sec:proximals}:

\begin{eqnarray*}
\Y^\T &\gets& \left(\Z\Z^\T+\mathbf{I}\right)^{-1}\left(\Z\X^\T+\W^\T + \tfrac{1}{\rho}\left(\Z\alpha^\T_\X-\alpha^\T_\Y\right)\right)\\
\Z &\gets& \left(\Y^\T\Y+\mathbf{I}\right)^{-1}\left(\Y^\T\X+\H + \tfrac{1}{\rho}\left(\Y^\T\alpha_\X-\alpha_\Z\right)\right)\\
\X &\gets& \frac{\left(\rho\Y\Z-\alpha_\X-\1\right)+\sqrt{\left(\rho\Y\Z-\alpha_\X-\1\right)^2+4\rho\V}}{2\rho}\\
\W &\gets& \left(\Y+\tfrac{1}{\rho}\alpha_\Y\right)_+\\
\H &\gets& \left(\Z+\tfrac{1}{\rho}\alpha_\Z\right)_+.
\end{eqnarray*}

Note that the primal updates require solving linear systems of sizes $r \times r$, but that the overal complexity remains $O(rmn)$ per iteration (the same as multiplicative updates).\\

The updates for the dual variables $\alpha_\X$, $\alpha_\Y$ and $\alpha_\Z$ are then:

\begin{eqnarray*}
\alpha_\X &\gets& \alpha_\X + \rho\left(\X-\Y\Z\right)\\
\alpha_\Y &\gets& \alpha_\Y + \rho\left(\Y-\W\right)\\
\alpha_\Z &\gets& \alpha_\Z + \rho\left(\Z-\H\right).
\end{eqnarray*}

This formulation introduces a regularization parameter, $\rho\in\R_+$, that needs to be tuned (in our experiments we choose the best performing one from several candidates).\\

Our approach has the following differences: (1) we aim at solving alternatively \emph{convex} problems with a few steps of primal-dual algorithms for convex problems, as opposed to aiming at solving directly the non-convex problem with an iterative approach, (2) for the convex decomposition problem, we have certificates of guarantees, which can be of used for online methods for which decomposition problems are repeatedly solved~\citep{lefevre2011} and (3) we use a different splitting method, namely the one of~\citet{chambolle2011}, which does not require matrix inversions, and which allows us to compute all step-sizes in a data-driven way.


\section{Proposed Method}

In this section we  present a  formulation of the convex KL decomposition problem as a  first-order primal-dual algorithm (FPA), followed by the proposed NMF technique.


\subsection{Primal and dual computation}\label{sec:pd}

We consider a vector $a\in\R^p_+$ and a matrix $K\in\R^{p\times q}_+$ as known parameters, and $x\in\R^q_+$ as an unknown vector to be estimated, where the following expression holds,

\begin{equation*}
a\approx Kx,
\end{equation*}
and we aim at minimizing the KL divergence between $a$ and $Kx$.\\

This is equivalent to a ND problem as defined in Problems (\ref{prob:nd1}) and (\ref{prob:nd2}), considering $a$ as a column of the given data, $K$ as the fixed factor, and $x$ as a column of the estimated factor, i.e., in \pref{prob:nd1} $a$ and $x$ are column vectors of $\V^\T$ and $\W^\T$ with the same index and $K$ is $\H^\T$, and in \pref{prob:nd2} $a$ and $x$ are columns of $\V$ and $\H$ with the same index and $K$ is $\W$.\\

The \emph{convex} ND problem with KL divergence is thus

\begin{equation}
\underset{x\in\R^q_+}{\operatorname{minimize}}\; - \sum_{i=1}^pa_i\left(\log(K_ix/a_i) + 1\right) + \sum_{i=1}^pK_ix,
\label{prob:primal}
\end{equation}

which may be written as

\begin{equation}
\label{eq:AA}
\begin{aligned}
& \underset{x\in\mathcal{X}}{\operatorname{minimize}}
& & F(Kx) + G(x),
\end{aligned}
\end{equation}

with
\begin{eqnarray*}
F(z) &=& - \sum_{i=1}^pa_i\left(\log(z_i/a_i) + 1\right) \\
G(x) &=& \mathbbm{1}_{x\succeq0} + \sum_{i=1}^pK_ix.
\end{eqnarray*}

Following \citet{pock2009,chambolle2011}, we obtain the dual problem

\begin{equation*}
\begin{aligned}
& \underset{y\in\mathcal{Y}}{\operatorname{maximize}}
& & -F^\ast(y) - G^\ast(-K^\ast y),
\end{aligned}
\end{equation*}

with 
\begin{eqnarray*}
F^\ast(y) & = & \sup_z\left\{y^\T z - F(z)\right\} =  -\sum_{i=1}^pa_i\log\left(-y_i\right) \\
G^\ast(y) & = &  \sup_{x}\left\{y^\T x - G(x)\right\} = \mathbbm{1}_{y\preceq K^\T\1}.
\end{eqnarray*}

We then get the dual problem

\begin{equation}
\begin{aligned}
& \underset{K^\T(-y) \;\preceq\; K^\T\1}{\operatorname{maximize}}
& & a^\T\log\left(-y\right).
\end{aligned}
\label{prob:dual}
\end{equation}

In order to provide a certificate of optimality, we have to make sure that the constraint $K^\T(-y) \;\preceq\; K^\T\1$ is satisfied. Therefore, when it is not satisfied, we project as follows:

\begin{equation*}
y \gets y/\max\{K^\T(-y)\oslash K^\T\1\},
\end{equation*}

where $\oslash$ represents the entry-wise division operator.


\subsection{Primal-dual algorithm}
\label{sec:proximals}

The general FPA framework of the approach proposed by \citeauthor{chambolle2011} for \pref{eq:AA} is presented in \autoref{algo:1}.

\RestyleAlgo{ruled}
\begin{algorithm}[htbp]
\caption{First-order primal-dual algorithm.}\label{algo:1}
\vspace{3mm}
Select $K\in\R^{p\times q}_+$, $x\in\R^q_+$,  $\sigma>0$, and $\tau>0$\;\vspace{1mm}

Set $\bar{x}=x_{old}=x$, and $y=Kx$\;\vspace{3mm}

\For{$N$ iterations}{\vspace{1mm}

$y \gets \prox_{\sigma F^\ast}(y-\sigma K\bar{x})$\;

$x \gets \prox_{\tau G}(x-\tau K^\ast y)$\;

$\bar{x} \gets 2x - x_{old}$\;

$x_{old} \gets x$\;\vspace{1mm}

}\vspace{3mm}

\Return{$x^\star = x.$}\vspace{3mm}
\end{algorithm}

\autoref{algo:1} requires the computation of  proximal operators $\prox_{\sigma F^\ast}(y)$ and $\prox_{\tau G}(x)$. These are defined as follows:

\begin{eqnarray*}
\prox_{\sigma F^\ast}(y) &=& \arg\min_v\left\{ \frac{\|v-y\|^2}{2\sigma} + F^\ast(v)\right\},\;\mathrm{and}\\
\prox_{\tau G}(x) &=& \arg\min_u\left\{ \frac{\|u-x\|^2}{2\tau} + G(u)\right\}.
\end{eqnarray*}

For further details, see \citep{boyd2004,rockafellar1997}.\\

Using the convex functions $F^\ast$ and $G$, we can easily solve the problems for the proximal operators and derive the following closed-form solution operators

\begin{eqnarray*}
\prox_{\sigma F^\ast}(y) &=& \frac{1}{2}\left(y - \sqrt{y\circ y + 4\sigma a}\right),\;\mathrm{and}\\
\prox_{\tau G}(x) &=& \left( x - \tau K^\T\1 \right)_+.
\end{eqnarray*}
The detailed derivation of these operators may be found in the \app{app:prox}, the first one was already computed by~\citet{sun2014}.


\subsection{Automatic heuristic selection of $\sigma$ and $\tau$}\label{sec:stepsizes}
 
In this section, we provide a heuristic way to select   $\sigma$ and $\tau$ without user intervention, based on the convergence result of~\citet[Theorem 1]{chambolle2011}, which states that (a) the step-sizes have to satisfy $\tau\sigma\|K\|^2 \leq 1$, where $\|K\|=\max\{\|Kx\|:x\in\mathcal{X}\; \textit{with}\; \|x\|\leq1\}$ is the largest singular value of $K$; and (b) the convergence rate is controlled by the quantity

$$C = \frac{\|y_0-y^\star\|^2}{2\sigma} + \frac{\|x_0-x^\star\|^2}{2\tau},$$ 
 
where $(x^\star,y^\star)$ is an optimal primal/dual pair. If $(x^\star,y^\star)$ was known, we could thus consider the following minimization problem with the constraint $\tau\sigma\|K\|^2 \leq 1$:
  
\begin{eqnarray*}
&& \min_{\sigma,\tau} \frac{\|y_0-y^\star\|^2}{2\sigma} + \frac{\|x_0-x^\star\|^2}{2\tau} \\
&\iff& \min_\sigma \frac{\|y_0-y^\star\|^2}{2\sigma} + \frac{\|x_0-x^\star\|^2}{2}\sigma\|K\|^2.
\end{eqnarray*}

Applying first order conditions, we find that

\begin{eqnarray*}
\sigma = \frac{\|y_0-y^\star\|}{\|x_0-x^\star\|}\frac{1}{\|K\|} &\mathrm{and}& \tau = \frac{\|x_0-x^\star\|}{\|y_0-y^\star\|}\frac{1}{\|K\|}.
\end{eqnarray*}

However, we do not know the  optimal pair $(x^\star,y^\star)$ and we use heuristic replacements. That is, we consider the unknown constants $\alpha$ and $\beta$, and assume that $x^\star = \alpha\1$ and $y^\star = \beta\1$ solve Problems (\ref{prob:primal}) and (\ref{prob:dual}). Letting $(x_0,y_0) = (\mathbf{0},\mathbf{0})$ we have

\begin{eqnarray*}
\|x_0-x^\star\| = |\alpha|\sqrt{q} &\mathrm{and}& \|y_0-y^\star\| = |\beta|\sqrt{p}.
\end{eqnarray*}

Plugging $x^\star$ to \pref{prob:primal}, we are able to find that $\alpha = \frac{\1^\T a}{\1^\T K\1}>0$. Now, using optimality conditions: $y^\star\circ(Kx^\star) = -a$, we obtain $\beta = -1$.\\

The updated version of the parameters is:

\begin{eqnarray*}
\sigma = \sqrt{\frac{p}{q}}\frac{1}{\alpha\|K\|} &\mathrm{and}& \tau = \sqrt{\frac{q}{p}}\frac{\alpha}{\|K\|}.
\end{eqnarray*}

Finally, an automatic heuristic selection of step sizes $\sigma$ and $\tau$ is as follows:

\begin{eqnarray*}
\sigma = \frac{\sqrt{p}\sum_{i=1}^pK_i\1}{\sqrt{q}\|K\|\sum_{i=1}^pa_i} &\mathrm{and}& \tau = \frac{\sqrt{q}\sum_{i=1}^pa_i}{\sqrt{p}\|K\|\sum_{i=1}^pK_i\1}.
\end{eqnarray*}

Note the invariance by rescaling of $a$ and $K$.


\subsection{Implementation}

The proposed method is based on \autoref{algo:1}. The required parameters to solve each ND problem are

\begin{multicols}{2}
\begin{itemize}
\item \pref{prob:nd1}:
\begin{itemize}
\item $a\; \gets \left(\V^\T\right)_i$
\item $K \gets \H^\T$
\item $x\; \gets \left(\W^\T\right)_i$
\item $\sigma \gets \sqrt{\frac{m}{r}}\frac{\1^\T\H\1}{\1^\T\left(\V^\T\right)_i\|\H\|}$
\item $\tau \gets \sqrt{\frac{r}{m}}\frac{\1^\T\left(\V^\T\right)_i}{\1^\T\H\1\|\H\|}$
\end{itemize}
\item \pref{prob:nd2}:
\begin{itemize}
\item $a\; \gets \V_i$
\item $K \gets \W$
\item $x\; \gets \H_i$
\item $\sigma \gets \sqrt{\frac{n}{r}}\frac{\1^\T\W\1}{\1^\T\V_i\|\W\|}$
\item $\tau \gets \sqrt{\frac{r}{n}}\frac{\1^\T\V_i}{\1^\T\W\1\|\W\|}$.
\end{itemize}
\end{itemize}
\end{multicols}

The previous summary treats each ND problem by columns. For algorithmic efficiency, we work directly with the matrices, e.g., $a\in\R^{n\times m}_+$ instead of $\R^n_+$.
We also include normalization steps such that the columns of $\W$ have sums equal to $1$.
The stopping criteria is enabled for maximum number of iterations (access to data) and for duality gap tolerance. 


\subsection{Extension to NMF}\label{sec:implementation}

A pseudo-code of the first-order primal-dual algorithm for non-negative matrix factorization can be found in \autoref{algo:2}. It corresponds to alternating between minimizing with respect to $\H$ and minimizing with respect to $\W$. A key algorithmic choice is the number of inner iterations \texttt{iter}$_{ND}$ of the convex method, which we consider in \sref{sec:results}.\\

The running-time complexity is still $O(rnm)$ for each inner iterations. Note moreover, that computing the largest singular value of $\H$ or $\W$ (required for the heuristic selection of step-sizes everytime we switch from one convex problem to the other) is of order $O(r \max\{m,n\})$ and is thus negligible compared to the iteration cost.


\RestyleAlgo{ruled}
\begin{algorithm}[ht!]
\caption{Proposed technique.}\label{algo:2}
\vspace{3mm}
Select $\V\in\R^{n\times m}_+$, $\W_0\in\R^{n\times r}_+$, and $\H_0\in\R^{r\times m}_+$\;\vspace{3mm}

Set $\W = \bar{\W} = \W_{old} = \W_0$, $\H = \bar{\H} = \H_{old} = \H_0$, and $\chi = \W\H$\;\vspace{3mm}

\While{stopping criteria not reached}{\vspace{3mm}

Normalize $\W$ and set $\sigma=\sqrt{\frac{m}{r}}\frac{\1^\T\H\1}{\1^\T\V^\T\|\H\|}\1$, $\tau=\sqrt{\frac{r}{m}}\frac{\1^\T\V^\T}{\1^\T\H\1\|\H\|}\1$, and $\H(-\chi^\T) \leq \H\1$\;\vspace{3mm}

\For{\texttt{iter}$_{ND}$ iterations}{\vspace{3mm}

$\chi^\T \gets \chi^\T - \sigma\circ\left(\bar{\W}\H\right)^\T$\;

$\chi^\T \gets \frac{1}{2}\left(\chi^\T - \sqrt{\chi^\T\circ\chi^\T + 4\sigma\circ\V^\T}\right)$\;

$\W^\T \gets \left( \W^\T - \tau \circ \left(\H\left(\chi^\T + \1 \right) \right) \right)_+$\;

$\bar{\W}^\T \gets 2\W^\T - \W_{old}^\T$\;

$\W_{old}^\T \gets \W^\T$\;\vspace{3mm}

}\vspace{3mm}

Normalize $\H$ and set $\sigma=\sqrt{\frac{n}{r}}\frac{\1^\T\W\1}{\1^\T\V\|\W\|}\1$, $\tau=\sqrt{\frac{r}{n}}\frac{\1^\T\V}{\1^\T\W\1\|\W\|}\1$, and $\W^\T(-\chi) \leq \W^\T\1$\;\vspace{3mm}

\For{\texttt{iter}$_{ND}$ iterations}{\vspace{3mm}

$\chi \gets \chi - \sigma \circ \left(\W\bar{\H}\right)$\;

$\chi \gets \frac{1}{2}\left(\chi - \sqrt{\chi \circ \chi + 4\sigma \circ \V}\right)$\;

$\H \gets \left( \H - \tau \circ \left(\W^\T\left(\chi + \1\right) \right) \right)_+$\;

$\bar{\H} \gets 2\H - \H_{old}$\;

$\H_{old} \gets \H$\;\vspace{3mm}

}\vspace{3mm}

}\vspace{3mm}

\Return{$\W^\star = \W$, and $\H^\star = \H$.}\vspace{3mm}

\end{algorithm}


%
\subsection{Extension to topic models}
Probabilistic latent semantic analysis~\citep{hofmann1999probabilistic} or latent Dirichlet allocation \citep{blei2003}, generative probabilistic models for collections of discrete data, have been extensively used in text analysis. Their formulations are related to ours in \pref{prob:primal}, where we just need to include an additional constraint: $\1^\T x=1$. In this sense, if we modify $G$, i.e., $G(x) = \mathbbm{1}\{\1^\T x=1;\;x\succeq0\} + 1^\T Kx$, we can use \autoref{algo:1} to find the latent topics. It is important to mention that herein $\prox_{\tau G}(x)$ does not have a closed solution, but can be efficiently solved with dedicated methods for orthogonal projections on the simplex~\citep{maculan1989linear}.


\section{Experimental Results}\label{sec:results}

The proposed technique was tested on synthetic data, the CBCL face images database and a music excerpt from a recognized jazz song by Louis Armstrong \& His Hot Five. The performance of the proposed first-order primal-dual algorithm (FPA) was compared against the traditional multiplicative updates algorithm (MUA) by \citet{lee2000} and the contemporary alternating direction method of multipliers (ADMM) by \citet{sun2014}. The three techniques were implemented in Matlab.


\subsection{Synthetic data}

A given matrix $\V$ of size $n=200$ and $m=1000$ is randomly generated from the uniform distribution $\mathcal{U}(0,750)$. The low-rank element was set to $r=15$. Initializations $\W_0$ and $\H_0$ are defined by the standard normal distribution's magnitude plus a small offset.

\begin{figure*}[h!]
\begin{center}
\begin{subfigure}[b]{3.2in}
\centering\includegraphics[width=\textwidth]{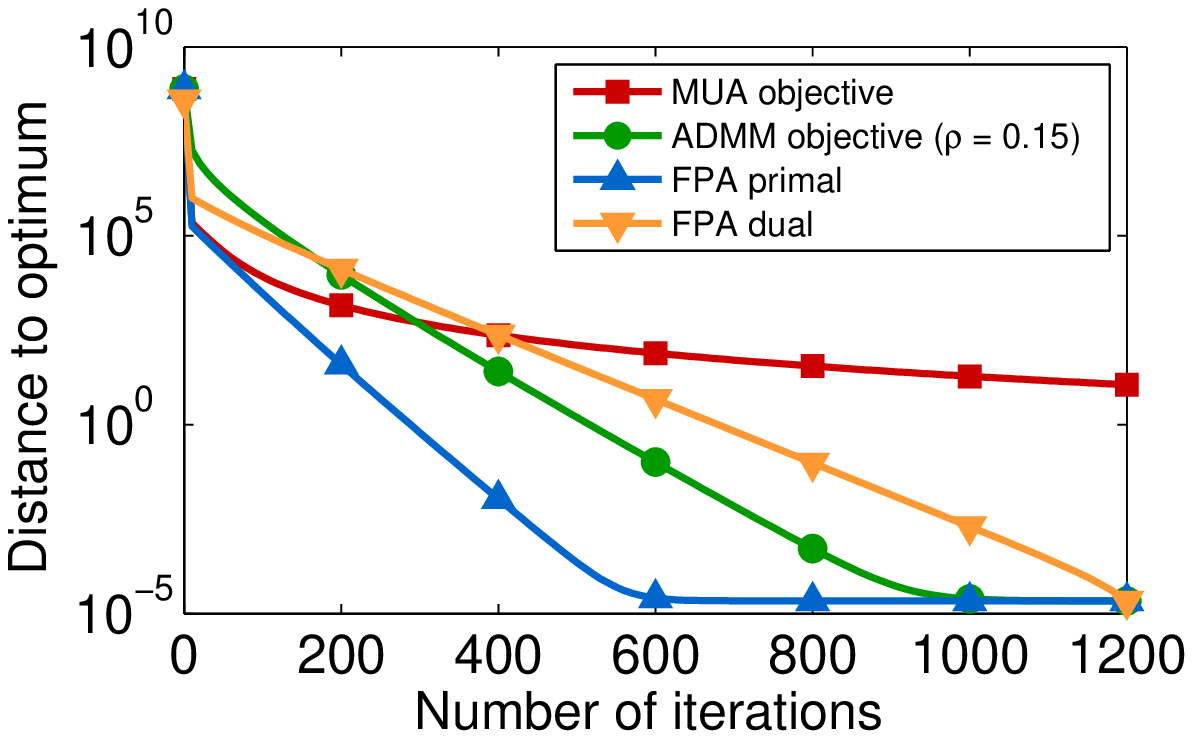}
\caption{Estimate $\W$ given $\H^\star$.} 
\vspace{3mm}
\end{subfigure}
\begin{subfigure}[b]{3.2in}
\centering\includegraphics[width=\textwidth]{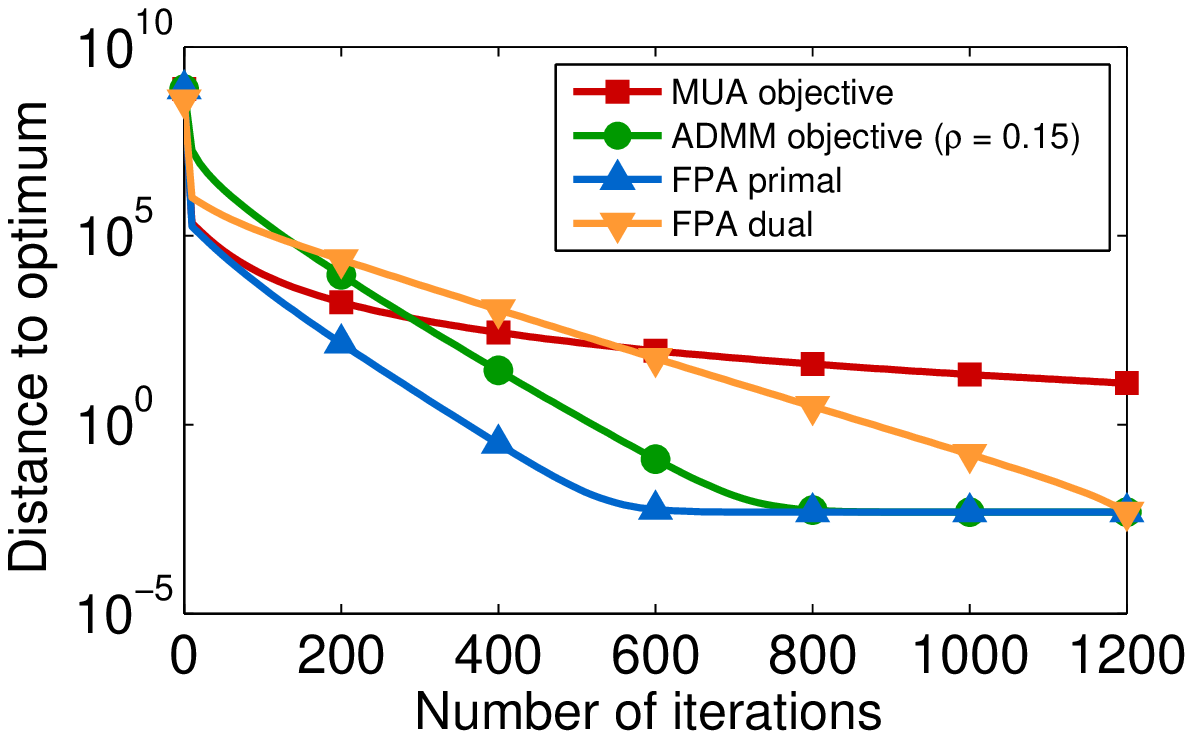}
\caption{Estimate $\H$ given $\W^\star$.}
\vspace{3mm}
\end{subfigure}
\begin{subfigure}[b]{3.2in}
\centering\includegraphics[width=\textwidth]{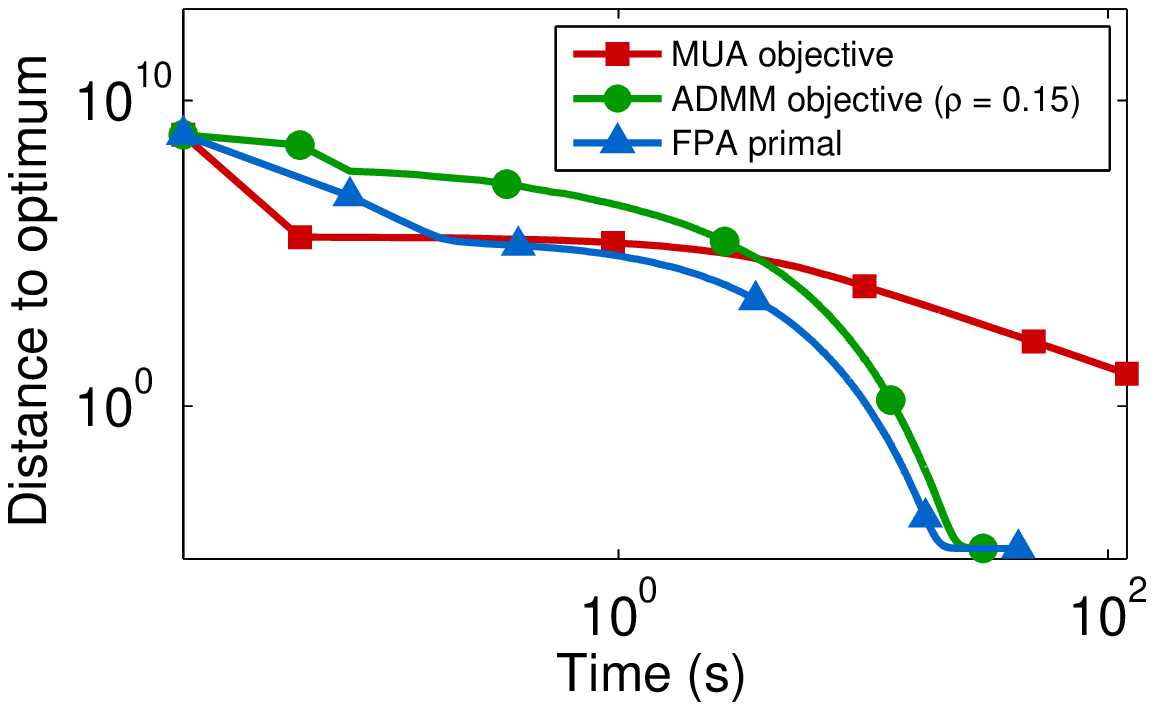}
\caption{Estimate $\W$ given $\H^\star$.}
\end{subfigure}
\begin{subfigure}[b]{3.2in}
\centering\includegraphics[width=\textwidth]{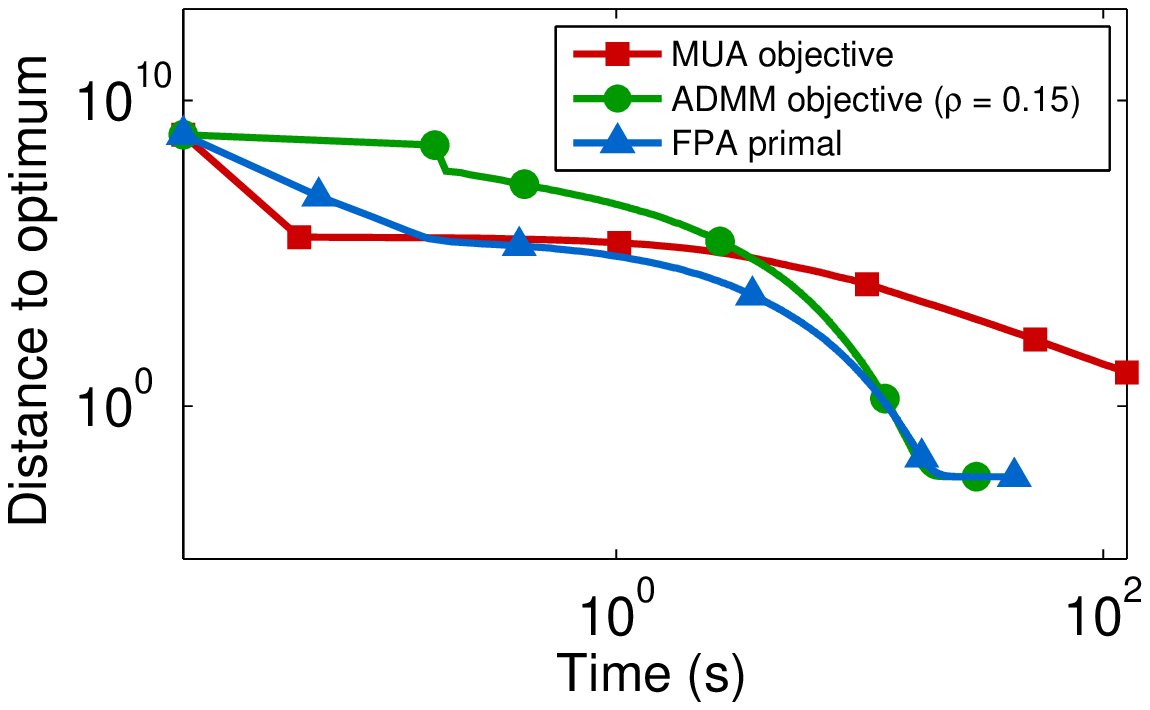}
\caption{Estimate $\H$ given $\W^\star$.}
\end{subfigure}
\end{center}
\caption{ND on synthetic data. (a-b) Distance to optimum versus iteration number. Distance to optimum reveals the difference between the values of the objective function and optimal point, $p^\star$. In the case of the dual function values, the distance to optimum is the difference between $p^\star$ and the dual points. (c-d) Distance to optimum versus time.}
\label{fig:synthetic1}
\end{figure*}

\subsubsection{ND problem}

To examine the accuracy of our method, we first apply \autoref{algo:2} to convex ND problems for fixed values of $n$, $m$ and $r$, solving separately Problems~(\ref{prob:nd1})~and~(\ref{prob:nd2}). For both problems, we set the number of iterations of the traditional MUA and contemporary ADMM to 1200, as well as for the proposed FPA.
Optimal factors $\W^\star$ and $\H^\star$ are obtained by running 5000 iterations of the MUA, starting from $\W_0$ and $\H_0$. For the first ND problem, we fix $\H$ to $\H^\star$ and estimate $\W$ starting from $\W_0$; for the second one, we fix $\W$ to $\W^\star$ and estimate $\H$ from $\H_0$. The optimal regularization parameter of  ADMM, the tuning parameter that controls the convergence rate, is $\rho=0.15$ (small values imply larger step sizes, which may result in faster convergence but also instability).
%
%
%
\frefs{fig:synthetic1}{a}{b} present us the distance to optimum of  MUA and ADMM, as well as for the primal and dual of our technique that reveals strong duality. The FPA and ADMM algorithm converge to the same point, whereas the MUA exhibits slow convergence. Note moreover the significant advantage towards our algorithm FPA, together with the fact that we set automatically all step-sizes.
In \frefs{fig:synthetic1}{c}{d}, we illustrate the distance to optimum versus time of the three methods.

\subsubsection{NMF problem}

The setting is slightly different as in the ND experiment, we increased the problem dimension to $n=250$, $m=2000$ and $r=50$, and repeat both previously presented experiments. For all methods, we set the number of iterations to 3000. The parameter \texttt{iter}$_{ND}$ indicates the number of iterations to solve each ND problem. We set \texttt{iter}$_{ND}$ to 5 iterations. To have a fair comparison between algorithms, for FPA, the number of iterations means access to data, i.e., we use 5 iterations to solve \pref{prob:nd1}  (as well as for \pref{prob:nd2}), and repeat this 600 times. The optimal regularization parameter of the ADMM is here $\rho=1$. 

\begin{figure*}[htbp]
\begin{center}
\begin{subfigure}[b]{3.2in}
\centering\includegraphics[width=\textwidth]{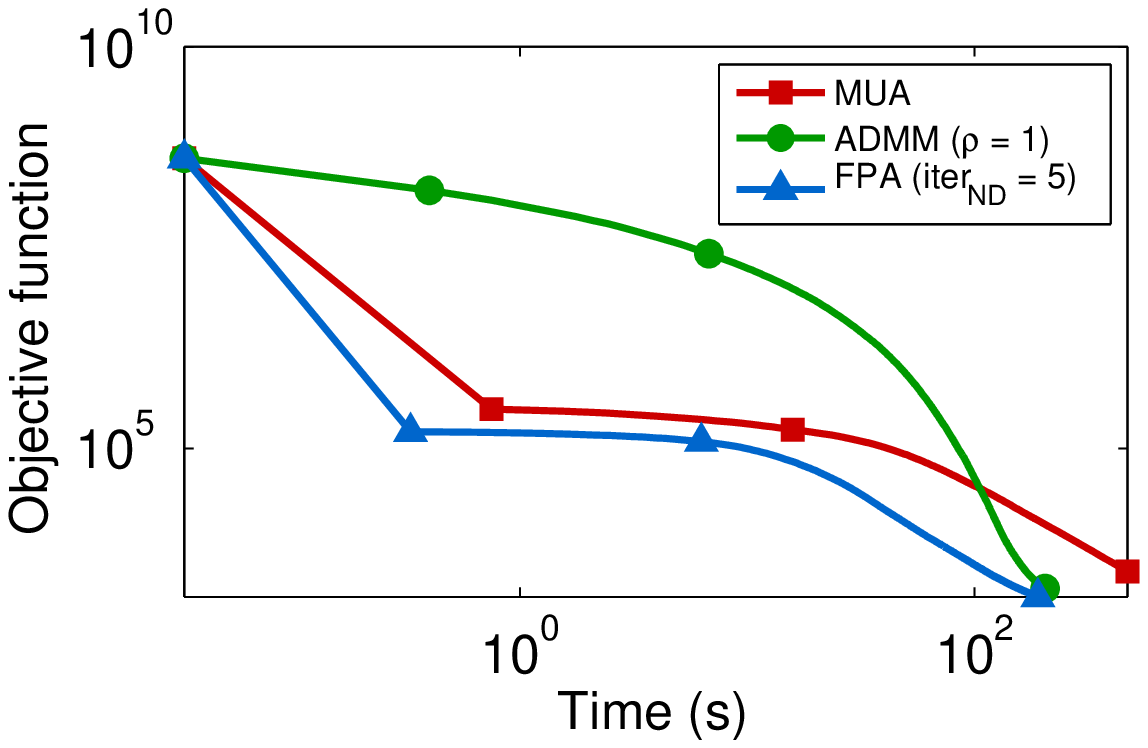}
\caption{Objective function versus time.}
\end{subfigure}
\begin{subfigure}[b]{3.2in}
\centering\includegraphics[width=\textwidth]{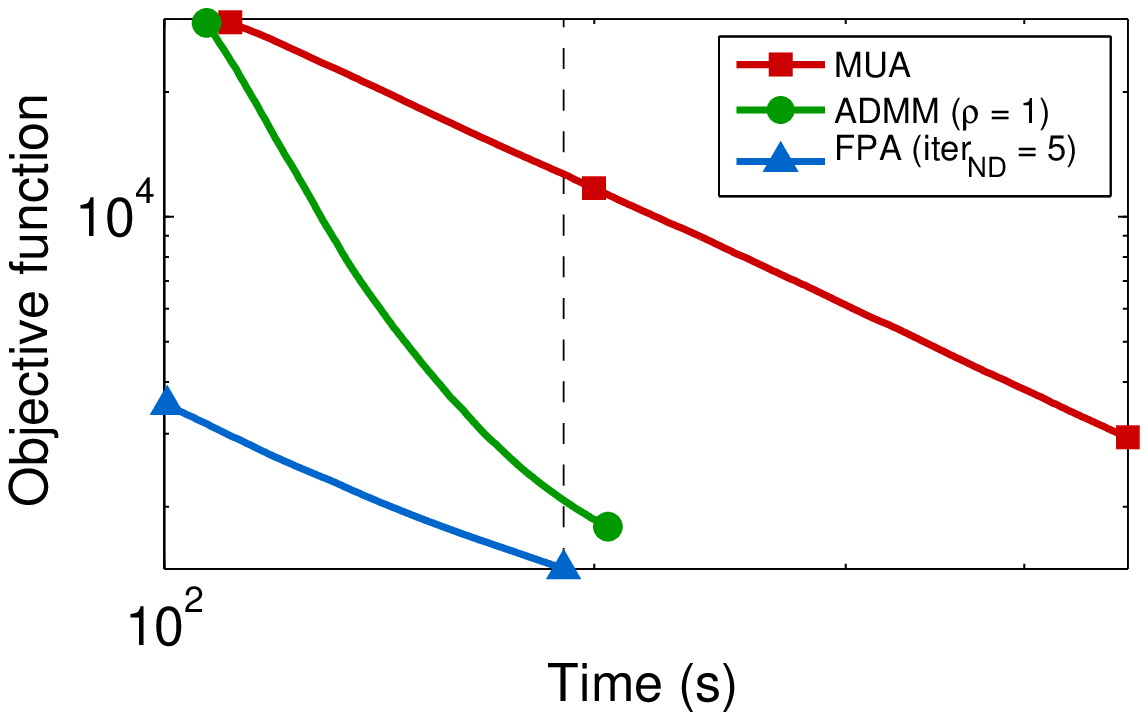}
\caption{Objective function versus time (zoomed).}
\end{subfigure}
\end{center}
\caption{
NMF on synthetic data. Recall that the dual function is not presented due to the non-convexity of the NMF problem.
} 
\label{fig:synthetic2}
\end{figure*}

In \autoref{fig:synthetic2} we present the objective function of the three algorithm for the non-convex \pref{prob:nmf}. The MUA initially reports high decrement in the objective, but as time increases it exhibits evident slow tail convergence. The FPA primal objective decreases dramatically in only seconds (few iterations), and furthermore, it presents fast tail convergence achieving the lowest objective value. The ADMM has poor initial performance, but then achieves an optimal value similar to the one obtained by FPA.
In order to show that   FPA converges faster and with lower computational cost, we store the cost function values and computation times at each iteration. The total time required by the FPA was 190s, whereas 205s by the ADMM and 473s by the MUA. Then we analyze the ADMM and MUA for the same time 190s (the vertical dotted line in \fref{fig:synthetic2}{b}), i.e., 2786 and 1211 iterations, respectively: the competitive algorithms have a significantly larger cost function for the same running time. The result of this experiment is illustrated in \fref{fig:synthetic2}{b}.
The results considering the objective function versus iteration number may be found in the \app{app:S}.

\subsubsection{NMF with warm restarts}

\begin{figure}[h!]
\centering\includegraphics[width=3.2in]{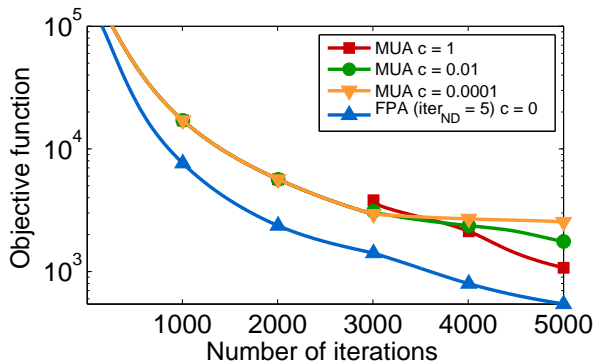}
\caption{NMF with warm restarts on synthetic data. Value of the objective function at each iteration.} 
\label{fig:synthetic3}
\end{figure}

The problem dimension is $n=150$, $m=2000$ and $r_1=50$. We run 3000 iterations of each method using initializations $\W_0$ and $\H_0$; then we increase ten times the low-rank element, $r_2=100$; and finally run 2000 more iterations, producing $\W_2$ and $\H_2$. The idea is to use as initializations the estimations obtained after the first 3000 iterations, $\W_1$ and $\H_1$, considering that the low-rank element changed. A trivial solution could be to include random entries so that $\W_1$ and $\H_1$ have the proper dimensions, but that increases the objective value, diminishing the estimations. On the other hand, if we include zero entries so that $\W_1$ and $\H_1$ have the proper dimensions, we would be in a saddle-point where none of the algorithms could perform. However, if we set only one factor with zero entries, $[\W_1,c\mathbf{1}]\in\R^{n\times r_2}$ with $c=0$, and the other one with non-zero values, $[\H_1;\nu]\in\R^{r_2\times m}$, we still maintain the same last objective value and perform  FPA. In this situation, MUA cannot perform either (because of the absorbing of zeros), therefore we try some values of $c$ to run the algorithm. \autoref{fig:synthetic3} illustrates the proposed experiment. Notice that as $c\to0$, the MUA starts to get stuck in a poor local optima, i.e., the one obtained with $\W_1$ and $\H_1$.  ADMM has a similar behavior as  FPA, therefore, it is not displayed.

\subsection{MIT-CBCL Face Database \#1}

\begin{figure}[htbp]
\centering\includegraphics*[scale=1]{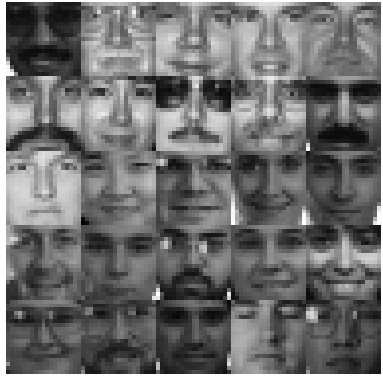}
\caption{MIT-CBCL Face Database \#1 samples.}
\label{fig:supp3}
\end{figure}

We use the CBCL face images database \citep{sung1996} composed of $m=2429$ images of size $n=361$ pixels. The low-rank element was set to $r=49$.
\autoref{fig:supp3} shows samples from the database.

\begin{figure*}[htbp]
\begin{center}
\begin{subfigure}[b]{3.2in}
\centering\includegraphics[width=\textwidth]{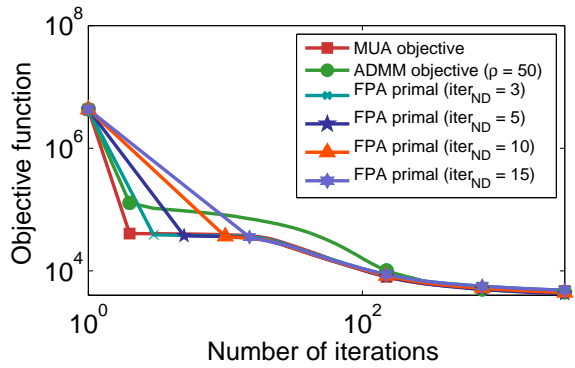}
\caption{Objective function versus iteration number.}
\end{subfigure}
\begin{subfigure}[b]{3.2in}
\centering\includegraphics[width=\textwidth]{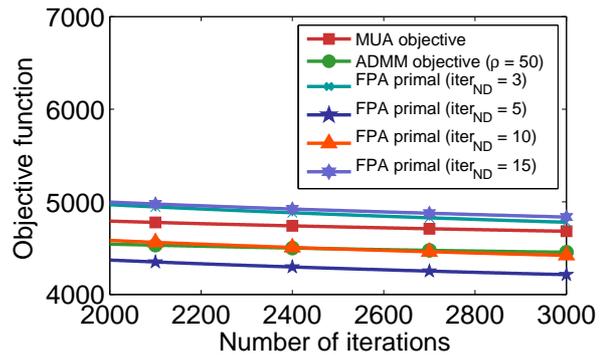}
\caption{Objective versus iteration number (zoomed).}
\end{subfigure}
\end{center}
\caption{NMF on the CBCL database. Value of the objective function at each iteration solving \pref{prob:nmf} varying the number of iterations to solve each ND problem.
} 
\label{fig:cbcl_obj}
\end{figure*}

Our next experiment is to determine the optimal the number of iterations for the current database. Therefore, we run 3000 iterations of  FPA, using 3, 5, 10 and 15 iterations for the ND problem. The MUA and ADMM ($\rho$=50) algorithms are performing as well. \autoref{fig:cbcl_obj} illustrates the decay of the objective function of the FPA, MUA and ADMM algorithms.

We appreciate that setting the number of iterations to 3 yield to over-alternation, whereas using 15 or even more iterations result in an under-alternating method. Using 10 iterations reveal good performance, but the best trade-off is obtained with 5 iterations. Therefore, we set \texttt{iter}$_{ND} = 5$, i.e., the number of iterations to solve \pref{prob:nd1} and \pref{prob:nd2}. All following results in the MIT-CBCL Face Database \#1 are with the same setting.\\ 

\begin{figure*}[h!]
\begin{center}
\begin{subfigure}[b]{3.2in}
\centering\includegraphics[width=\textwidth]{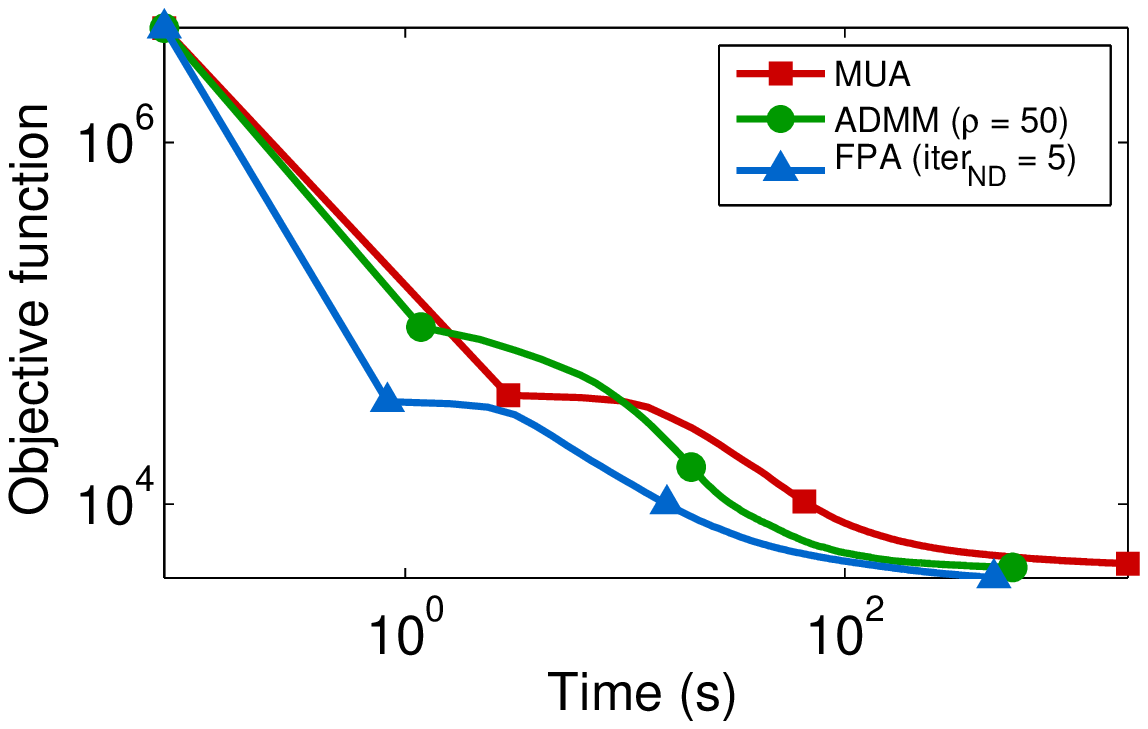}
\caption{Objective function versus time.}
\end{subfigure}
\begin{subfigure}[b]{3.2in}
\centering\includegraphics[width=\textwidth]{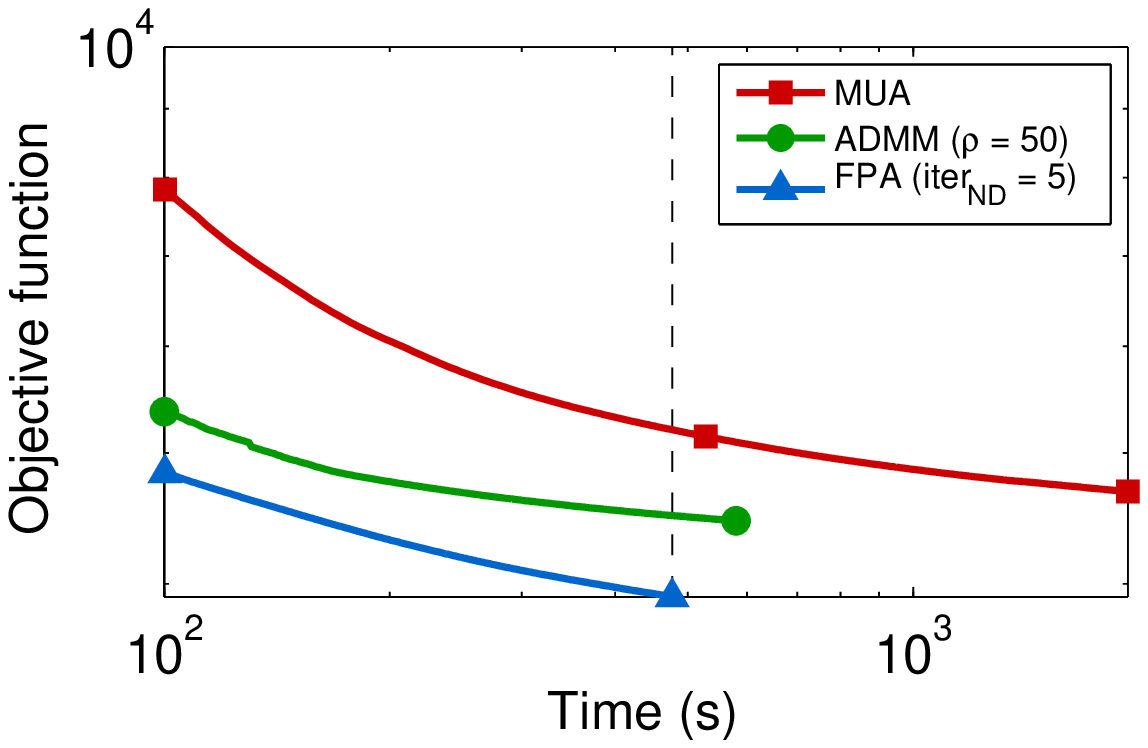}
\caption{Objective function versus time (zoomed).}
\end{subfigure}
\end{center}
\caption{NMF on the CBCL face image database.
} 
\label{fig:cbcl_nmf}
\end{figure*}

Finally, in \fref{fig:cbcl_nmf}{a} we present the objective function of the three algorithm for the non-convex \pref{prob:nmf}, where all algorithms perform  similarly. However, in the zoomed \fref{fig:cbcl_nmf}{b} we can appreciate that the MUA presents the slowest convergence, whereas the proposed method the fastest one.
The results considering the objective function versus iteration number may be found in the \app{app:F}.

\subsection{Music excerpt from the song ``My Heart (Will Always Lead Me Back to You)"}

\begin{figure}[htbp]
\centering\includegraphics[width=3.2in]{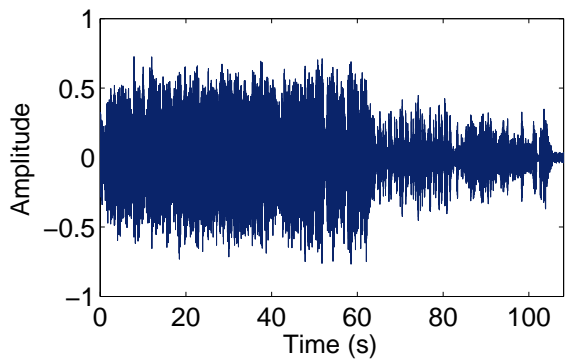}
\caption{Time-domain recorded signal.} 
\label{fig:music_signal}
\end{figure}

The last experiment is to decompose a 108-second-long music excerpt from ``My Heart (Will Always Lead Me Back to You)" by Louis Armstrong \& His Hot Five in the 1920s \citep{fevotte2009a}. The time-domain recorded signal is illustrated in \autoref{fig:music_signal}.

\begin{figure}[htbp]
\centering\includegraphics[width=3.2in]{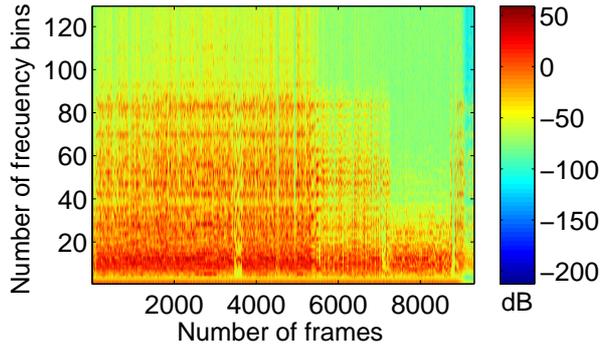}
\caption{Log-power spectrogram.} 
\label{fig:music_spectrogram}
\end{figure}

The recording consists of a trumpet, a double bass, a clarinet, a trombone, and a piano. The recorded signal is original unprocessed mono material contaminated with noise. The signal was downsampled to 11025 kHz, yielding 1,19$\times10^6$ samples. The Fourier Transform of the recorded signal was computed using a sinebell analysis window of length 23 ms with 50\% overlap between two frames, leading to $m=9312$ frames and $n=129$ frequency bins. Additionally, we set $r=10$. \autoref{fig:music_spectrogram} illustrates the previously described spectrogram.

The decomposition of the song is produced by the three algorithms. We initialize them with the same random values $\W_0$ and $\H_0$. For a fair competition, the number of iterations is set to 5000 for MUA and ADMM, and for our algorithm FPA we consider it as access to data, i.e., we use 5 iterations for the ND, repeating it 1000 times. For comparison, we measure the computation time of the three techniques. FPA has a run time of 13 min, whereas the ADMM ($\rho=10$) one of 15 min and the MUA one of 80 min. 

\begin{figure}[h!]
\begin{center}
\begin{subfigure}[b]{3.2in}
\centering\includegraphics[width=\textwidth]{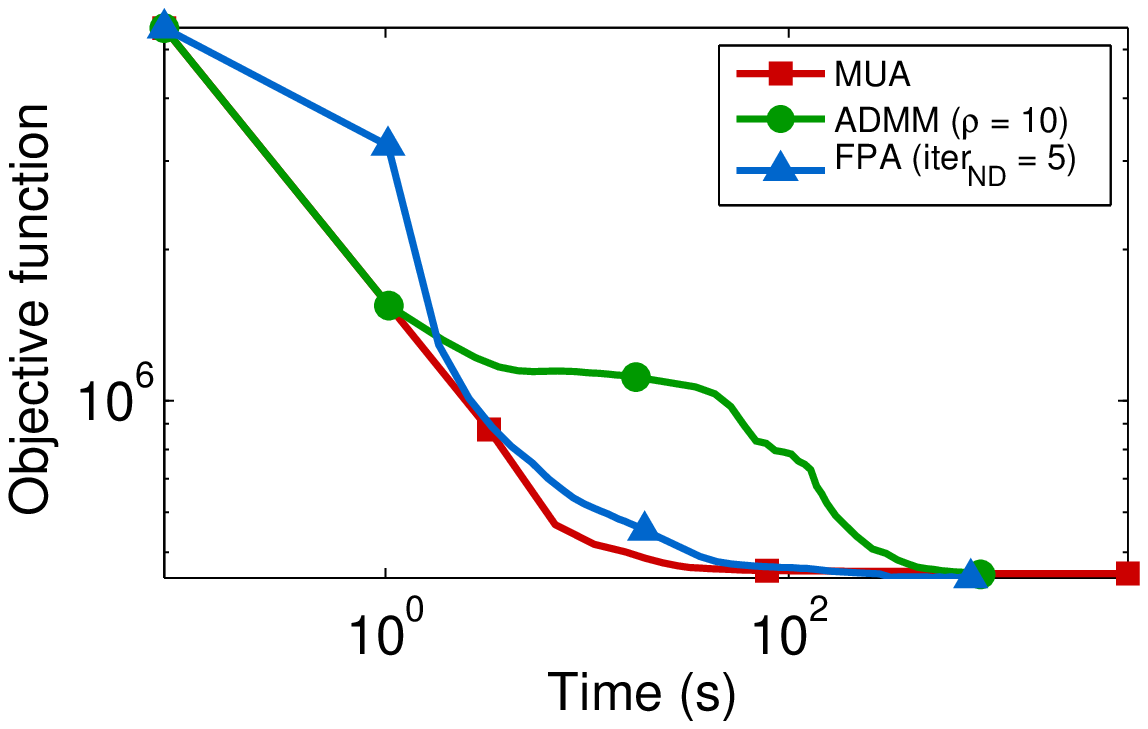}
\caption{Objective function versus time.}
\end{subfigure}
\begin{subfigure}[b]{3.2in}
\centering\includegraphics[width=\textwidth]{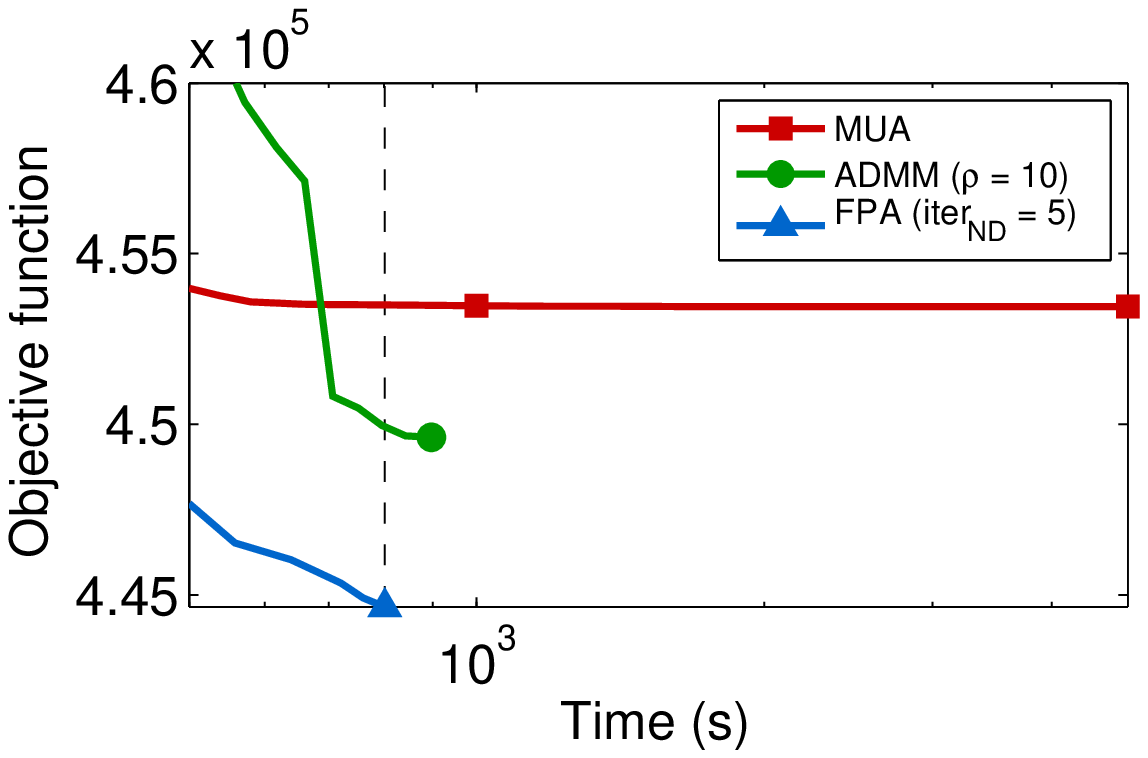}
\caption{Objective function versus time (zoomed).}
\end{subfigure}
\end{center}
\caption{NMF on an excerpt of Armstrong's song.
 } 
\label{fig:music_time}
\end{figure}

In this experiment, \autoref{fig:music_time} illustrates the evolution of the objective of the three techniques along \emph{time}. Initially the MUA obtained the lowest objective value, but as previously discussed, as the number of iterations increases the MUA starts exhibiting evident slow tail convergence and since approximately 100s it is reached by the FPA and shows no further substantial decrement, i.e., it gets stuck in a worse local optima. FPA converges to a slight lower cost value, overpassing   MUA. Finally,   ADMM reveals   a slow initial performance on this dataset, but later converges to a similar point as the previous algorithms.
The results considering the objective function versus iteration number may be found in the \app{app:M}.


\section{Conclusion}

We have presented an alternating projected gradient descent technique for NMF that minimizes the KL divergence loss; this approach solves convex ND problems with the FPA. Our approach demonstrated faster convergence than the traditional MUA by \citet{lee2000} and contemporary ADMM by \citet{sun2014}. The FPA introduces a new parameter, the number of iterations for each convex ND problem. Experiments reveal that the number of iterations is mostly bounded between 3 and 10 iterations, which leads to a trivial tuning by the user. Therefore, our algorithm affords reasonable simplicity, where further user manipulation is not required. Finally, an extension to latent Dirichlet allocation and probabilistic latent semantic indexing can be easily implemented using our proposed method, thus allowing to go beyond the potential slowness of the expectation-maximization (EM) algorithm.

%
\subsubsection*{Acknowledgements}

This work was partially supported by a grant from the European Research Council (ERC SIERRA 239993).

%


\bibliographystyle{plainnat}
\bibliography{nmf_refs}


\newpage
\section*{Appendix}

\subsection*{Derivation of proximal operators}\label{app:prox}

The definition of the proximal operator of $F^\ast$ and $G$, i.e., $(I+\sigma\partial F^\ast)^{-1}(y)$ and $(I+\tau\partial G)^{-1}(x)$, respectively, is as follows \citep{pock2009,chambolle2011}:

\begin{eqnarray*}
(I + \sigma\partial F^\ast)^{-1}(y) &=& \arg\min_v\left\{ \frac{\|v-y\|^2}{2\sigma} + F^\ast(v)\right\},\;\mathrm{and}\\
(I + \tau\partial G)^{-1}(x) &=& \arg\min_u\left\{ \frac{\|u-x\|^2}{2\tau} + G(u)\right\},
\end{eqnarray*}

where $\partial F^\ast$ and $\partial G$ are the subgradients of the convex functions $F^\ast$ and $G$.\\

To facilitate the computation of $(I+\sigma\partial F^\ast)^{-1}(y)$, we consider Moreau's identity

\begin{eqnarray*}
y &=& \left(I + \tau\partial F^\ast\right)^{-1}\left(y\right) + \sigma\left(I + \frac{1}{\sigma}\partial F\right)^{-1}\left(\frac{y}{\sigma}\right) \\
   &=& \prox_{\sigma F^\ast}(y) + \sigma\;\prox_{F/\sigma}(y/\sigma).
\end{eqnarray*}

Let us consider the variable $v\in\mathcal{Y}$, and using Moreau's identity, we can compute

\begin{eqnarray*}
\prox_{\sigma F^\ast}(y) &=& y - \sigma\;\prox_{F/\sigma}(y/\sigma) \\
&=& y - \sigma\arg\min_v\left\{ \frac{\sigma}{2}\left\|v-\frac{y}{\sigma}\right\|^2 + F(v)\right\} \\
&=& y - \sigma\arg\min_v\left\{ \frac{\sigma}{2}\left\|v-\frac{y}{\sigma}\right\|^2 - \sum_{i=1}^na_i\left(\log\left(\frac{v_i}{a_i}\right) + 1\right)\right\} \\
&=& y - \sigma\arg\min_v\left\{\sum_{i=1}^n\frac{\sigma}{2}\left(v_i^2-\frac{2v_iy_i}{\sigma} + \frac{y_i^2}{\sigma^2}\right) - a_i\left(\log\left(\frac{v_i}{a_i}\right) + 1\right)\right\} \\
&=& y - \sigma\arg\min_v\left\{ \sum_{i=1}^n\frac{\sigma}{2}v_i^2 - y_iv_i - a_i\log\left(v_i\right)\right\}.
\end{eqnarray*}

Applying first order conditions to obtain the minimum:

\begin{eqnarray*}
\frac{d}{dv_i}\left\{\frac{\sigma}{2}v_i^2 - y_iv_i - a_i\log\left(v_i\right)\right\}=0 &\implies& \sigma v_i - y_i - \frac{a_i}{v_i}=0 \\
 &\implies& \sigma v_i^2 - y_iv_i - a_i=0 \\
 &\implies& v_i = \frac{y_i \pm \sqrt{y_i^2 + 4\sigma a_i}}{2\sigma} \\
 &\implies& v = \frac{y + \sqrt{y\circ y + 4\sigma a}}{2\sigma},\;\mathrm{as}\;v\succ0.
\end{eqnarray*}

Finally, the proximal operator is as follows:

\begin{equation*}
\prox_{\sigma F^\ast}(y) = \frac{1}{2}\left(y - \sqrt{y\circ y + 4\sigma a}\right).
\end{equation*}

For the second proximal operator, we consider $u\in\mathcal{X}$ and compute $\prox_{\tau G}(x)$ as

\begin{eqnarray*}
\prox_{\tau G}(x) &=&  \arg \min_{ u \succeq 0 } \left \{ \frac{ \| u - x \| ^2 }{ 2\tau } + G(u) \right \} \\
&=& \arg \min_{ u \succeq 0 } \left \{ \frac{ \| u - x \| ^2 }{ 2\tau } + \sum_{i=1}^n K_i u \right \} \\
&=& \left( x - \tau K^\T\mathbf{1} \right)_+.
\end{eqnarray*}


\subsection*{Synthetic data: additional results}

\subsubsection*{NMF problem}\label{app:S}

\begin{figure*}[htbp]
\begin{center}
\begin{subfigure}[b]{3in}
\centering\includegraphics[width=\textwidth]{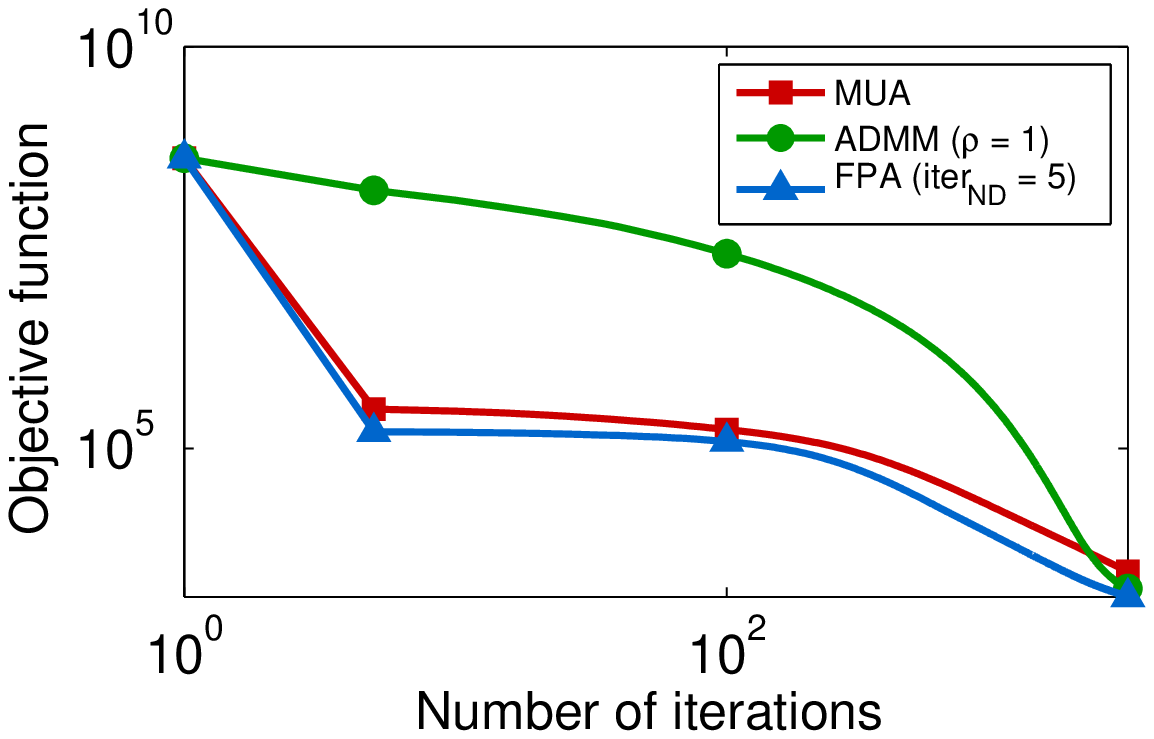}
\caption{Objective versus iteration number.}
\end{subfigure}
\begin{subfigure}[b]{3in}
\centering\includegraphics[width=\textwidth]{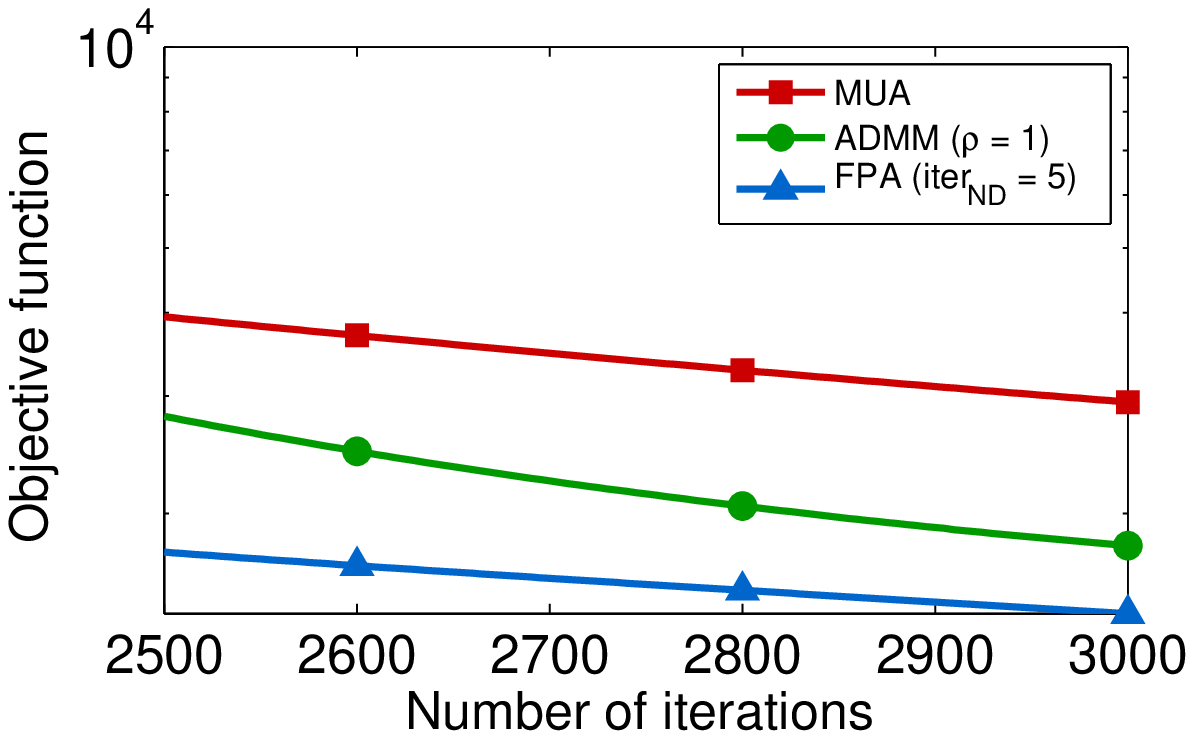}
\caption{Zoomed version.}
\end{subfigure}
\end{center}
\vspace{-5mm}
\caption{NMF on synthetic data. It is important to recall that the dual function is not presented due to the non-convexity of the NMF problem.}
\label{fig:supp2}
\end{figure*}

A given matrix $\V$ of size $n=250$ and $m=2000$ is randomly generated from the uniform distribution $\mathcal{U}(0,750)$. The low-rank element was set to $r=50$. For the three methods, we set the number of iterations to 3000. We set \texttt{iter}$_{ND}$ to 5 iterations. The optimal tuning parameter of the ADMM is here $\rho=1$. In \autoref{fig:supp2} we present the objective function versus iteration number of the three algorithms for the non-convex NMF problem. 

\subsection*{MIT-CBCL Face Database \#1: additional results}

\subsubsection*{ND problem}

\begin{figure*}[h!]
\begin{center}
\begin{subfigure}[b]{3in}
\centering\includegraphics[width=\textwidth]{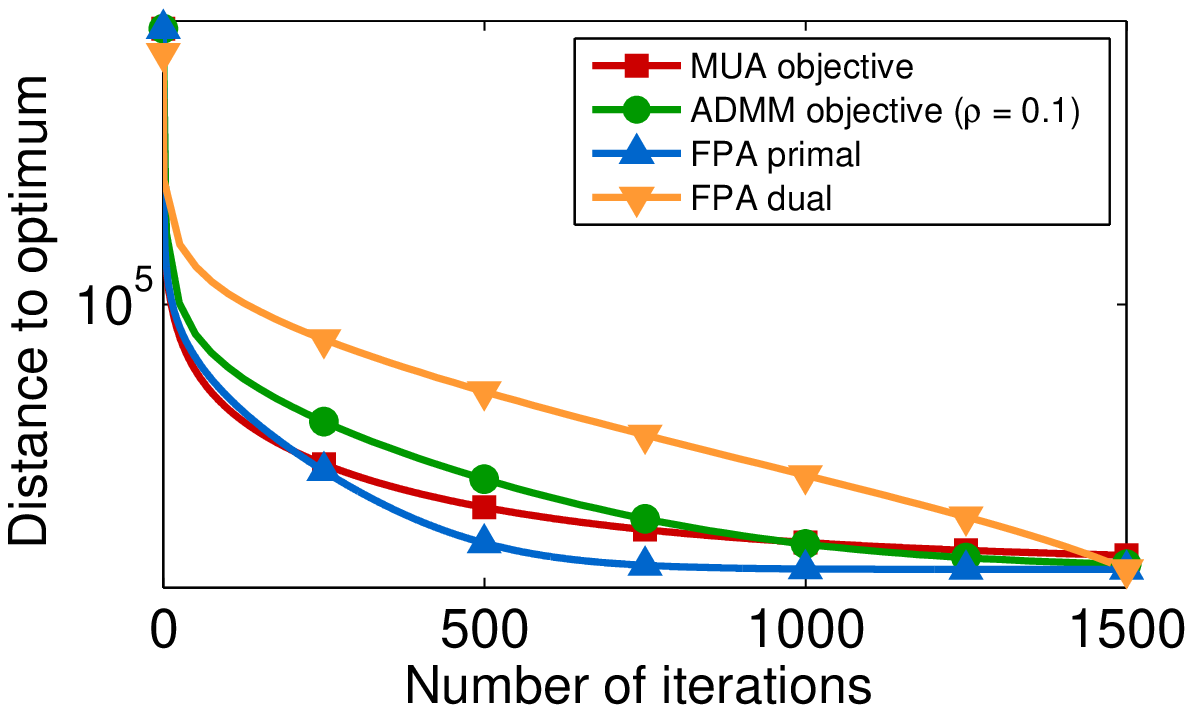}
\caption{Estimate $\W$ given $\H^\star$.} 
\end{subfigure}
\begin{subfigure}[b]{3in}
\centering\includegraphics[width=\textwidth]{figures/FigApp_F_ND_iter_fixedW.eps}
\caption{Estimate $\H$ given $\W^\star$.}
\end{subfigure}
\begin{subfigure}[b]{3in}
\centering\includegraphics[width=\textwidth]{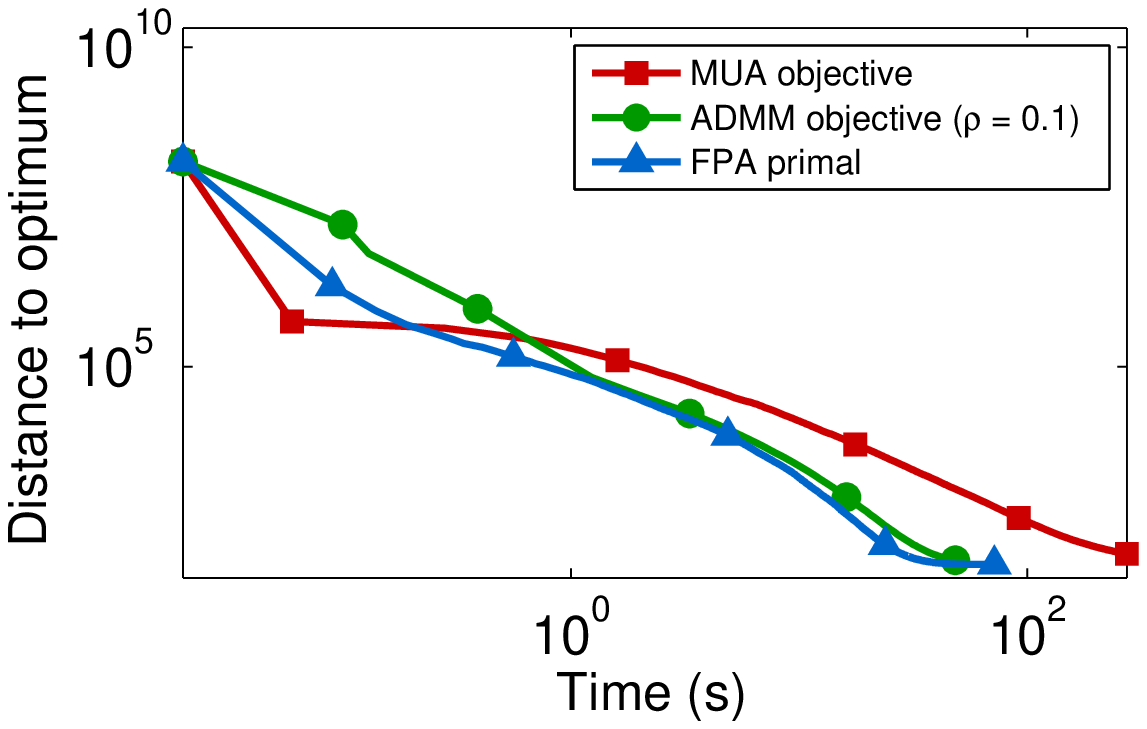}
\caption{Estimate $\W$ given $\H^\star$.}
\end{subfigure}
\begin{subfigure}[b]{3in}
\centering\includegraphics[width=\textwidth]{figures/FigApp_F_ND_time_fixedW.eps}
\caption{Estimate $\H$ given $\W^\star$.}
\end{subfigure}
\end{center}
\vspace{-5mm}
\caption{Distance to optimum versus (a-b) iteration number, and (c-d) time.} 
\label{fig:supp4}
\end{figure*}

We solve convex ND problems for fixed values of $n$, $m$ and $r$, setting the number of iterations of all algorithms to 1500.
Optimal factors $\W^\star$ and $\H^\star$ are obtained by running 5000 iterations of the MUA. The optimal tuning parameter of the ADMM is here $\rho=0.1$.
\frefs{fig:supp4}{a}{b} presents us the distance to optimum of the MUA and ADMM, as well as for the primal and dual of our technique that reveals strong duality in all experiments.
In \frefs{fig:supp4}{c}{d}, we illustrate the distance to optimum versus time of the three methods.

\subsubsection*{NMF problem}\label{app:F}

\begin{figure*}[h!]
\begin{center}
\begin{subfigure}[b]{3in}
\centering\includegraphics[width=\textwidth]{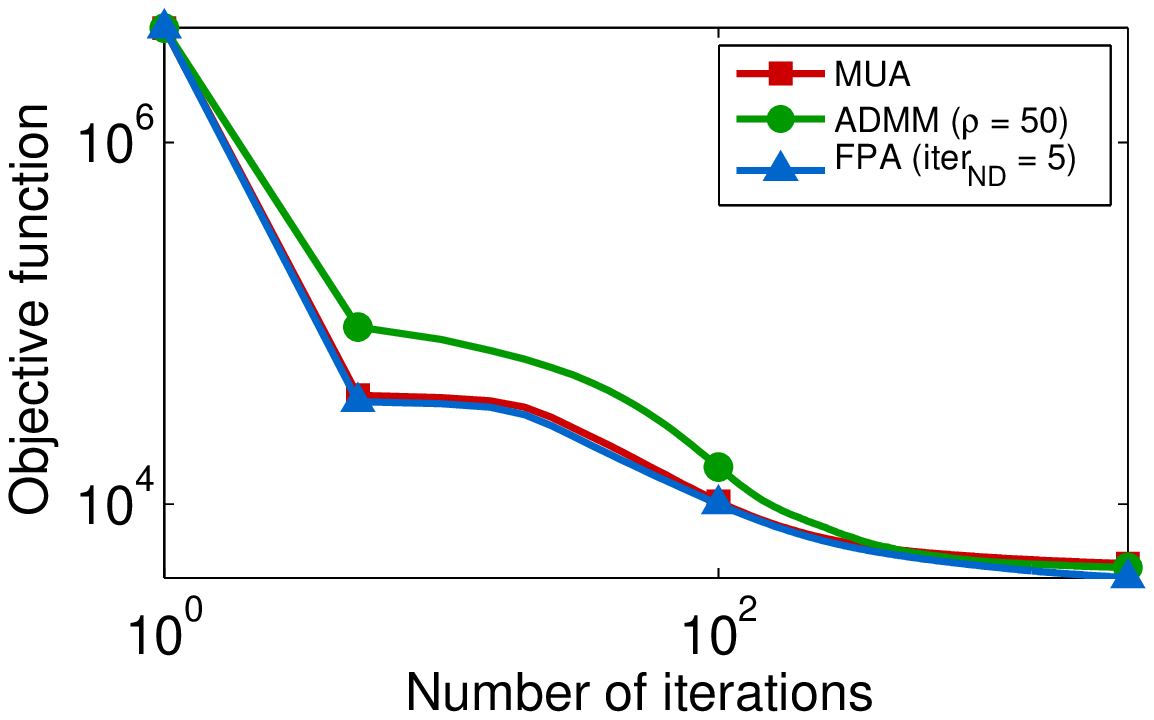}
\caption{Objective versus iteration number.}
\end{subfigure}
\begin{subfigure}[b]{3in}
\centering\includegraphics[width=\textwidth]{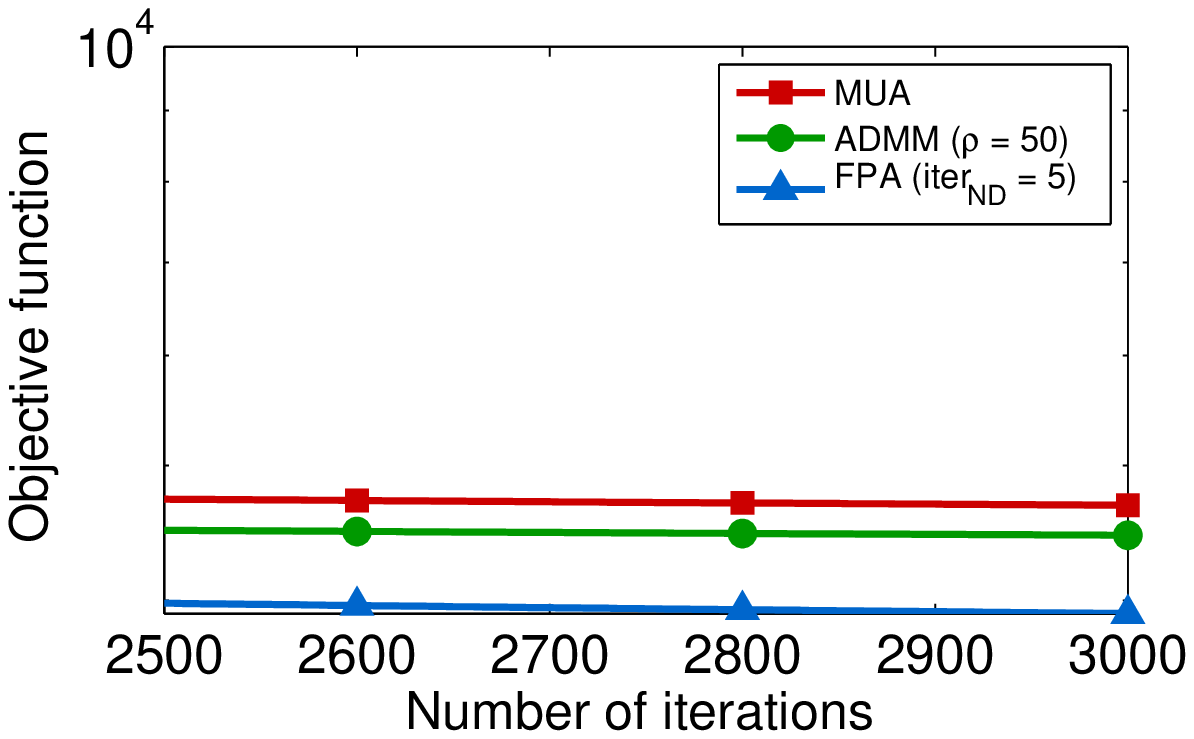}
\caption{Zoomed version.}
\end{subfigure}
\end{center}
\vspace{-5mm}
\caption{NMF on the CBCL face image database.}
\label{fig:supp5}
\end{figure*}

For all methods, we set the number of iterations to 3000. We set \texttt{iter}$_{ND}$ to 5 iterations. The optimal tuning parameter of the ADMM is here $\rho=50$. In \autoref{fig:supp5} we present the objective function versus iteration number of the three algorithms. 

\subsubsection*{Features learned from the CBCL face image database}

\begin{figure*}[h!]
\begin{center}
\begin{subfigure}[b]{2in}
\centering\includegraphics[width=\textwidth]{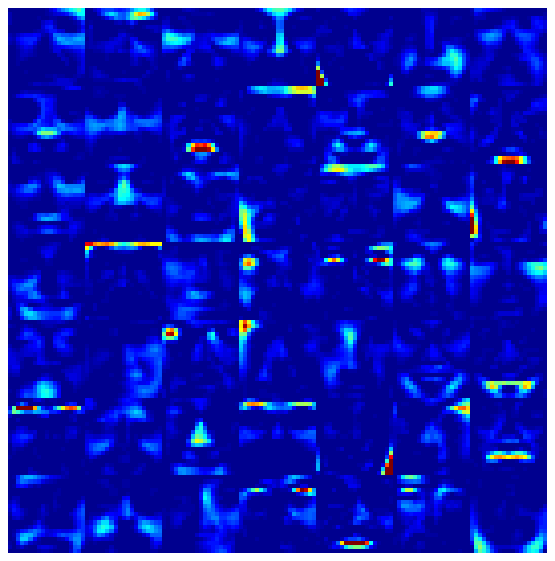}
\caption{MUA.}
\end{subfigure}
\begin{subfigure}[b]{2in}
\centering\includegraphics[width=\textwidth]{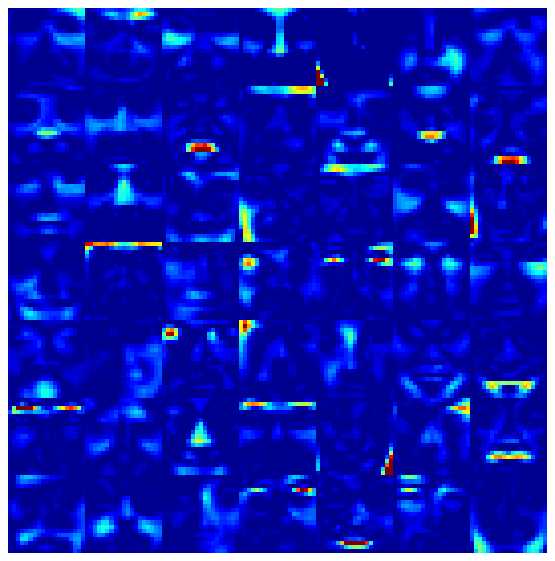}
\caption{ADMM.}
\end{subfigure}
\begin{subfigure}[b]{2in}
\centering\includegraphics[width=\textwidth]{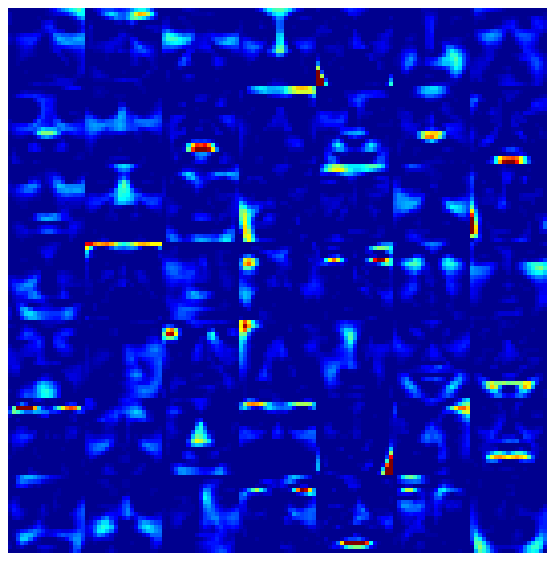}
\caption{FPA.}
\end{subfigure}
\end{center}
\vspace{-5mm}
\caption{Features learned from the CBCL face image database.} 
\label{fig:supp6}
\end{figure*}

The features learned from the CBCL face image database obtained with the three algorithms is presented in \autoref{fig:supp6}. The figure reveals the parts-based learned by the algorithm, i.e.~$\mathbf{W}$.

\subsection*{``My Heart (Will Always Lead Me Back to You)": additional results}

\begin{figure*}[h!]
\begin{center}
\begin{subfigure}[b]{3in}
\centering\includegraphics[width=\textwidth]{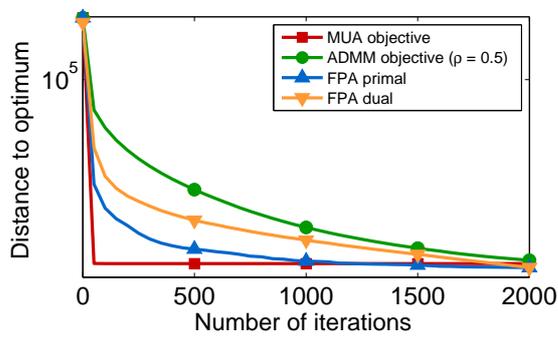}
\caption{Estimate $\W$ given $\H^\star$.} 
\end{subfigure}
\begin{subfigure}[b]{3in}
\centering\includegraphics[width=\textwidth]{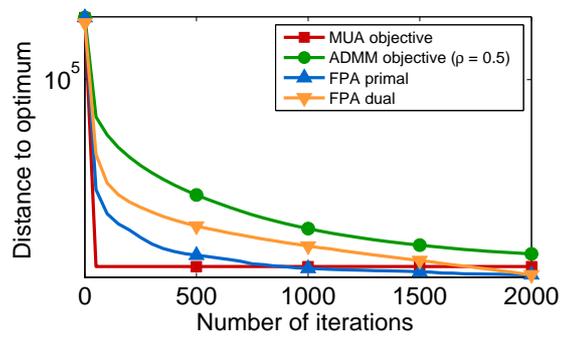}
\caption{Estimate $\H$ given $\W^\star$.}
\end{subfigure}
\begin{subfigure}[b]{3in}
\centering\includegraphics[width=\textwidth]{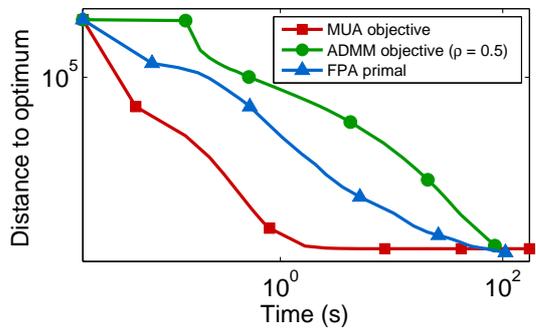}
\caption{Estimate $\W$ given $\H^\star$.}
\end{subfigure}
\begin{subfigure}[b]{3in}
\centering\includegraphics[width=\textwidth]{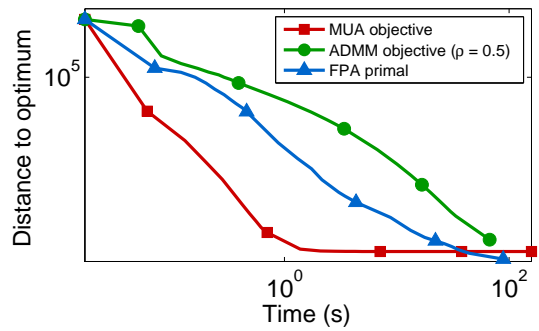}
\caption{Estimate $\H$ given $\W^\star$.}
\end{subfigure}
\end{center}
\vspace{-5mm}
\caption{Distance to optimum versus (a-b) iteration number, and (c-d) time.} 
\label{fig:supp9}
\end{figure*}

\subsubsection*{ND problem}

We solve convex ND problems for fixed values of $n$, $m$ and $r$, setting the number of iterations of all algorithms to 2000.
Optimal factors $\W^\star$ and $\H^\star$ are obtained by running 5000 iterations of the MUA. The optimal tuning parameter of the ADMM is here $\rho=0.5$.
\frefs{fig:supp9}{a}{b} presents us the distance to optimum of the MUA and ADMM, as well as for the primal and dual of our technique that reveals strong duality in all experiments.
In \frefs{fig:supp9}{c}{d}, we illustrate the distance to optimum versus time of the three methods.

\subsubsection*{NMF problem}\label{app:M}

\begin{figure*}[h!]
\begin{center}
\begin{subfigure}[b]{3in}
\centering\includegraphics[width=\textwidth]{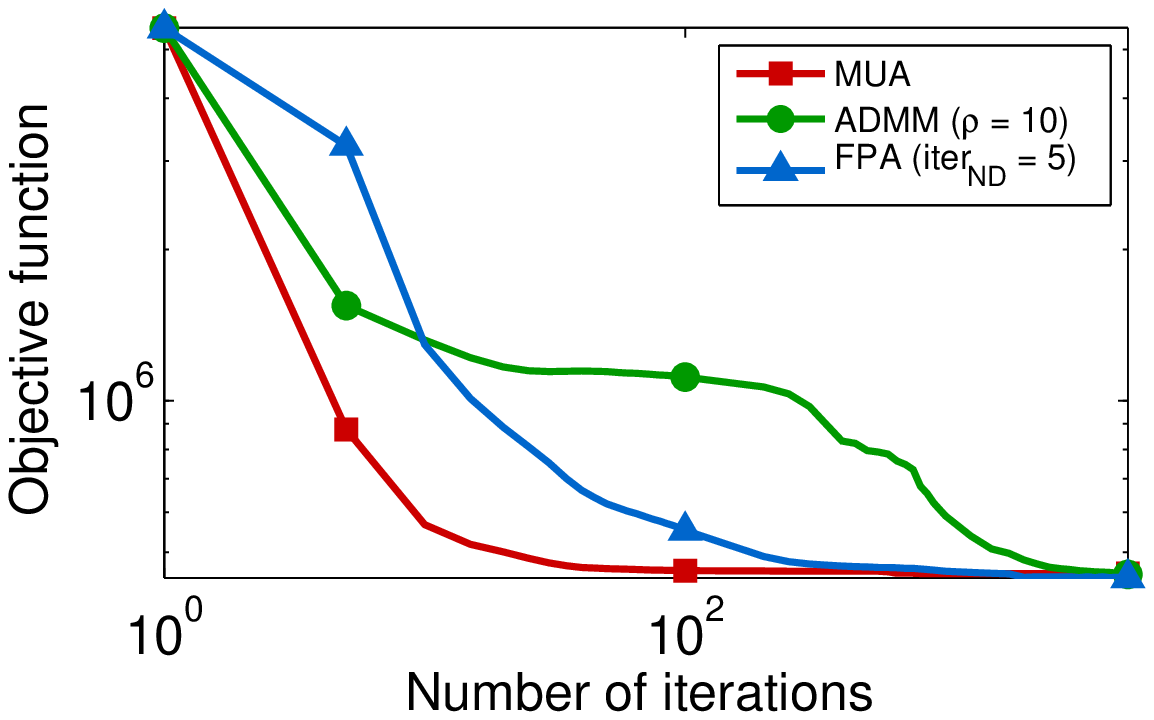}
\caption{Objective versus iteration number.}
\end{subfigure}
\begin{subfigure}[b]{3in}
\centering\includegraphics[width=\textwidth]{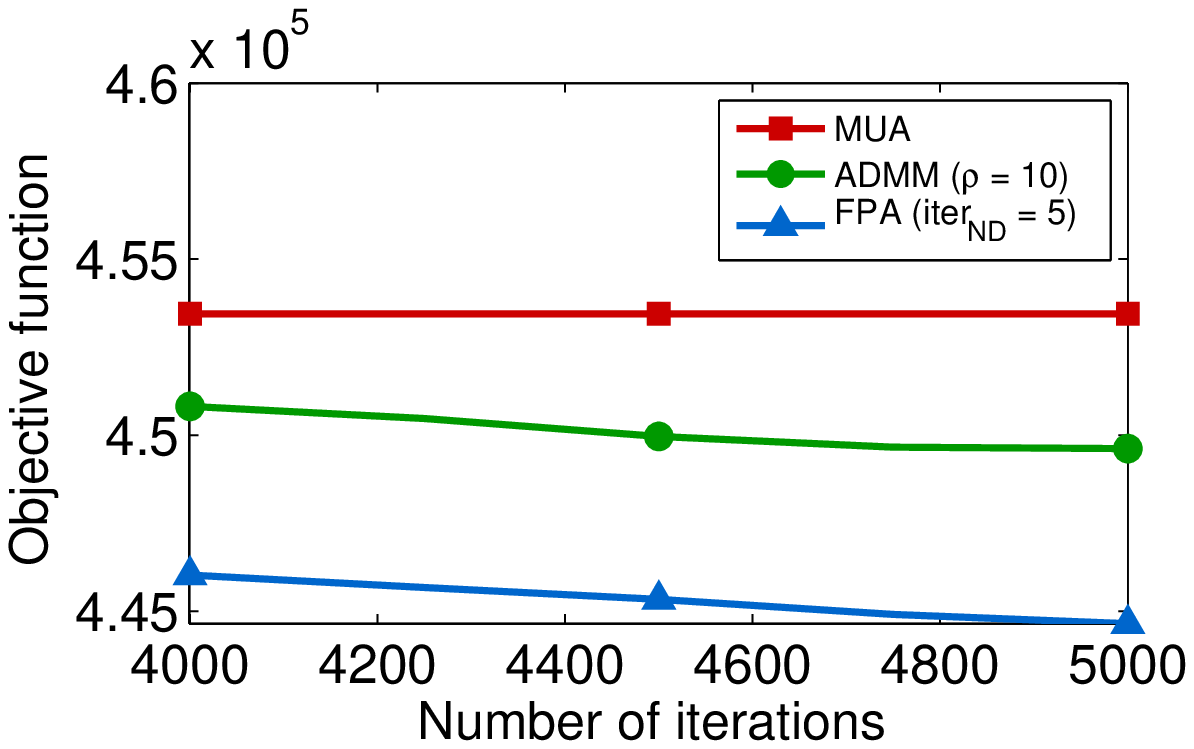}
\caption{Zoomed version.}
\end{subfigure}
\end{center}
\vspace{-5mm}
\caption{NMF on an excerpt of ArmstrongÕs song.}
\label{fig:supp10}
\end{figure*}

For all methods, we set the number of iterations to 5000. We set \texttt{iter}$_{ND}$ to 5 iterations. The optimal tuning parameter of the ADMM is here $\rho=10$. In \autoref{fig:supp10} we present the objective function versus iteration number of the three algorithms. 

\newpage

\subsubsection*{Features learned from the song}

\begin{figure*}[h!]
\begin{subfigure}[b]{0.48\textwidth}
\centering\includegraphics[scale=0.8]{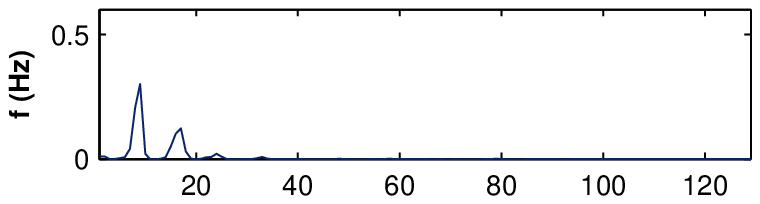}
\vspace{1mm}
\end{subfigure}
\hspace{4.75mm}
\begin{subfigure}[b]{0.48\textwidth}
\centering\includegraphics[scale=0.8]{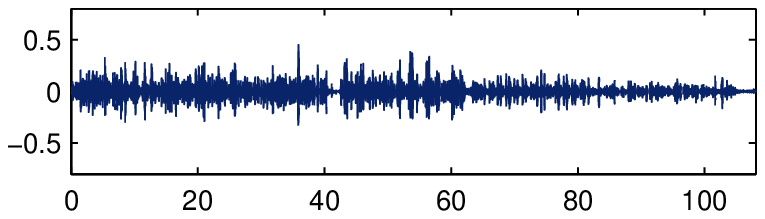}
\vspace{1mm}
\end{subfigure}
\begin{subfigure}[b]{0.48\textwidth}
\centering\includegraphics[scale=0.8]{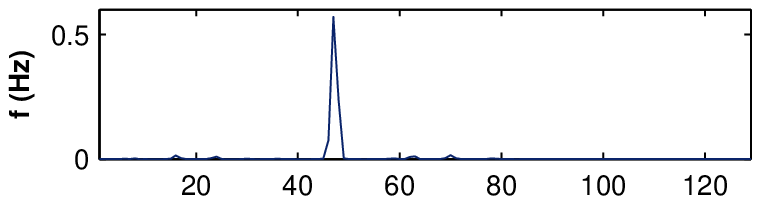}
\vspace{1mm}
\end{subfigure}
\hspace{4.75mm}
\begin{subfigure}[b]{0.48\textwidth}
\centering\includegraphics[scale=0.8]{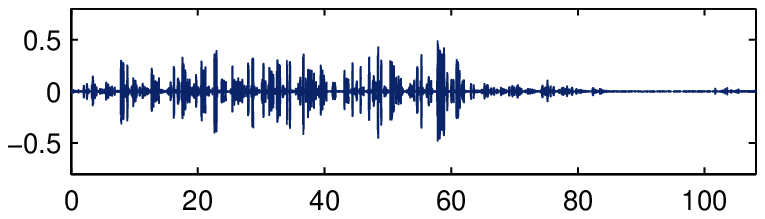}
\vspace{1mm}
\end{subfigure}
\begin{subfigure}[b]{0.48\textwidth}
\centering\includegraphics[scale=0.8]{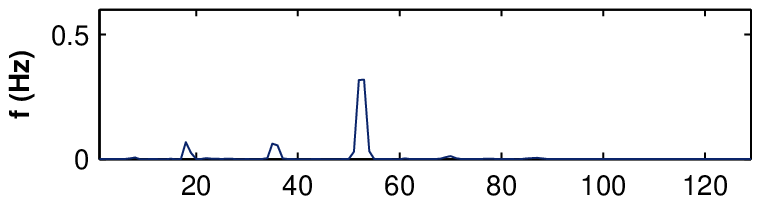}
\vspace{1mm}
\end{subfigure}
\hspace{4.75mm}
\begin{subfigure}[b]{0.48\textwidth}
\centering\includegraphics[scale=0.8]{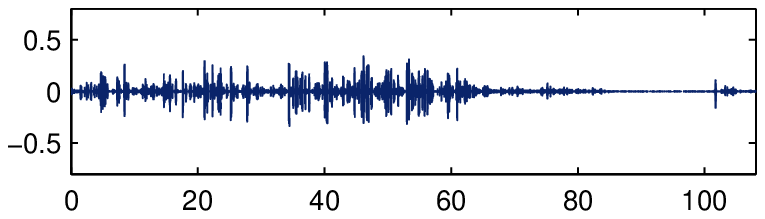}
\vspace{1mm}
\end{subfigure}
\begin{subfigure}[b]{0.48\textwidth}
\centering\includegraphics[scale=0.8]{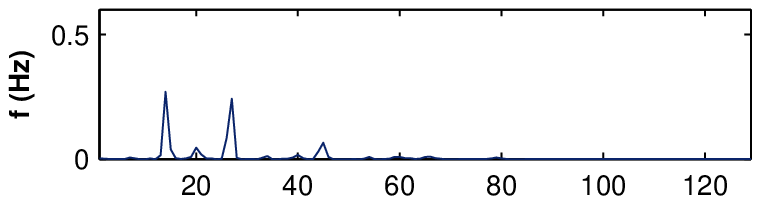}
\vspace{1mm}
\end{subfigure}
\hspace{4.75mm}
\begin{subfigure}[b]{0.48\textwidth}
\centering\includegraphics[scale=0.8]{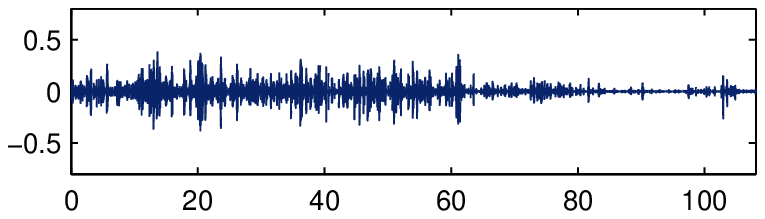}
\vspace{1mm}
\end{subfigure}
\begin{subfigure}[b]{0.48\textwidth}
\centering\includegraphics[scale=0.8]{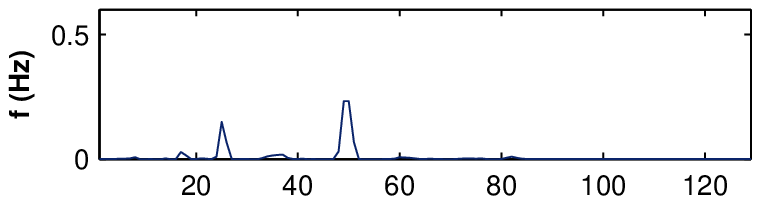}
\vspace{1mm}
\end{subfigure}
\hspace{4.75mm}
\begin{subfigure}[b]{0.48\textwidth}
\centering\includegraphics[scale=0.8]{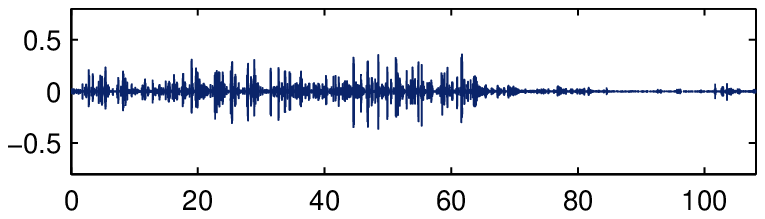} 
\vspace{1mm}
\end{subfigure}
\begin{subfigure}[b]{0.48\textwidth}
\centering\includegraphics[scale=0.8]{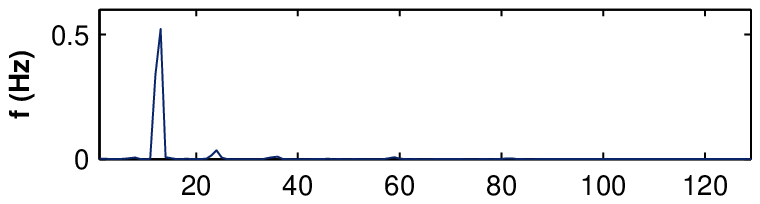}
\vspace{1mm}
\end{subfigure}
\hspace{4.75mm}
\begin{subfigure}[b]{0.48\textwidth}
\centering\includegraphics[scale=0.8]{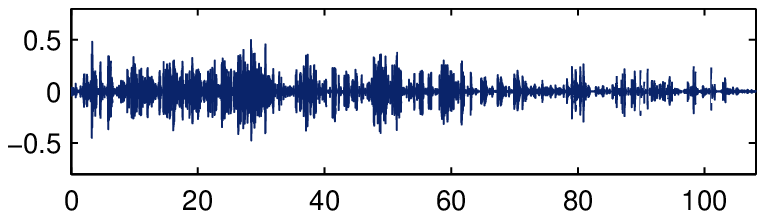} 
\vspace{1mm}
\end{subfigure}
\begin{subfigure}[b]{0.48\textwidth}
\centering\includegraphics[scale=0.8]{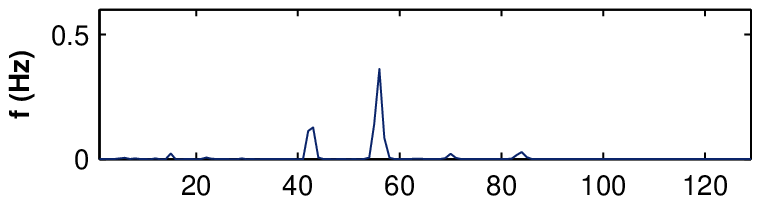}
\vspace{1mm}
\end{subfigure}
\hspace{4.75mm}
\begin{subfigure}[b]{0.48\textwidth}
\centering\includegraphics[scale=0.8]{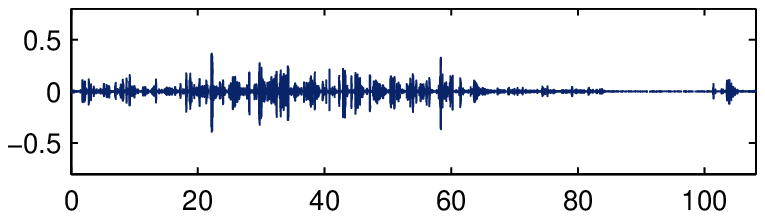}
\vspace{1mm}
\end{subfigure}
\begin{subfigure}[b]{0.48\textwidth}
\centering\includegraphics[scale=0.8]{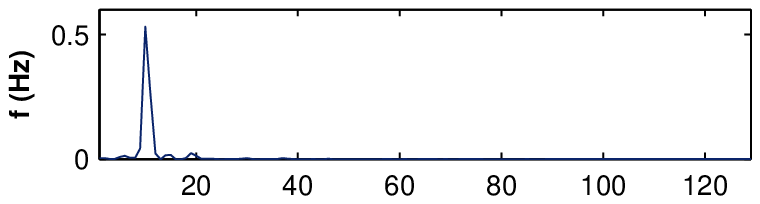}
\vspace{1mm}
\end{subfigure}
\hspace{4.75mm}
\begin{subfigure}[b]{0.48\textwidth}
\centering\includegraphics[scale=0.8]{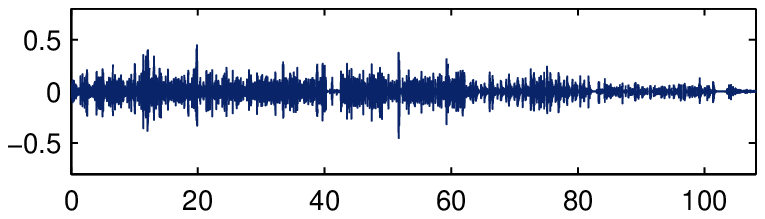}
\vspace{1mm}
\end{subfigure}
\begin{subfigure}[b]{0.48\textwidth}
\centering\includegraphics[scale=0.8]{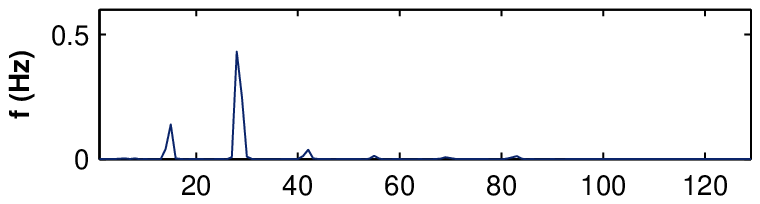}
\vspace{1mm}
\end{subfigure}
\hspace{4.75mm}
\begin{subfigure}[b]{0.48\textwidth}
\centering\includegraphics[scale=0.8]{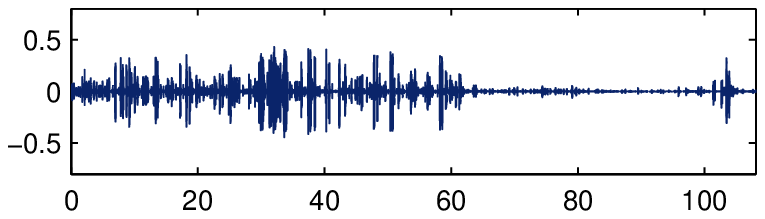}
\vspace{1mm}
\end{subfigure}
\begin{subfigure}[b]{0.48\textwidth}
\centering\includegraphics[scale=0.8]{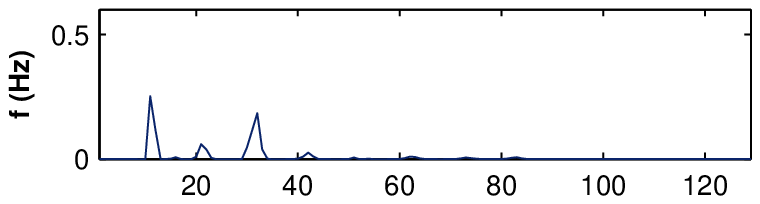}
\vspace{1mm}
\caption*{\hspace{0mm}\textbf{Frecuency bins}\\\vspace{3mm}\scriptsize{$\mathbf{W}$, parts-based learned by the FPA.}} 
\end{subfigure}
\hspace{4.75mm}
\begin{subfigure}[b]{0.48\textwidth}
\centering\includegraphics[scale=0.8]{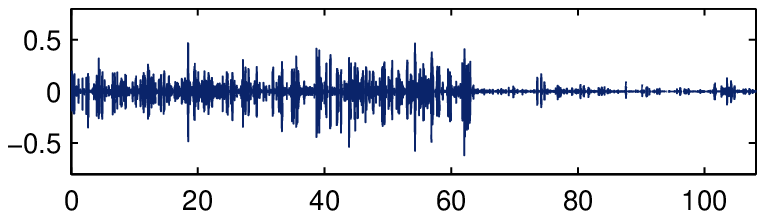}
\vspace{1mm}
\caption*{\hspace{3mm}\textbf{Time (s)}\\\vspace{3mm}\scriptsize{Time-domain recovered signal.}}
\end{subfigure}
\caption{Decomposition of Louis Armstrong and band song.} 
\label{fig:sound}
\end{figure*}

The decomposition of the song by Louis Armstrong and band obtained with the proposed FPA is presented in \autoref{fig:sound}, revealing the parts-based learned by the algorithm, i.e., $\mathbf{W}$. The time-domain signal is recovered from Wiener filtering \citep{fevotte2009a}. 

\end{document}